\documentclass[conference]{IEEEtran}
\IEEEoverridecommandlockouts
\usepackage{cite}
\usepackage{amsmath,amssymb,amsfonts}
\usepackage{algorithmic}
\usepackage{graphicx}
\usepackage{textcomp}
\usepackage{xcolor}
\usepackage{comment}
\usepackage{booktabs}
\usepackage{caption}
\usepackage{array}
\usepackage{multirow}
\usepackage{tabularx}
\usepackage{graphicx}
\usepackage{comment}
\usepackage{caption}
\usepackage{subcaption}

\usepackage{amsmath}

\def\BibTeX{{\rm B\kern-.05em{\sc i\kern-.025em b}\kern-.08em
    T\kern-.1667em\lower.7ex\hbox{E}\kern-.125emX}}

\begin{document}

\title{Activation Sparsity Opportunities for Compressing General Large Language Models\\
}


\author{\IEEEauthorblockN{Nobel Dhar, Bobin Deng, Md Romyull Islam, Kazi Fahim Ahmad Nasif, Liang Zhao, Kun Suo}
\IEEEauthorblockA{
College of Computing and Software Engineering, Kennesaw State University\\
Email:
ndhar@students.kennesaw.edu, bdeng2@kennesaw.edu, \{mislam22, knasif\}@students.kennesaw.edu,\\
\{lzhao10, ksuo\}@kennesaw.edu}
}

\maketitle

\begin{abstract}

Deploying local AI models, such as Large Language Models (LLMs), to edge devices can substantially enhance devices' independent capabilities, alleviate the server's burden, and lower the response time. Owing to these tremendous potentials, many big tech companies have been actively promoting edge LLM evolution and released several lightweight Small Language Models (SLMs) to bridge this gap. However, SLMs currently only work well on limited real-world applications. We still have huge motivations to deploy more powerful (larger-scale) AI models on edge devices and enhance their smartness level. Unlike the conventional approaches for AI model compression, we investigate from activation sparsity. The activation sparsity method is orthogonal and combinable with existing techniques to maximize compression rate while maintaining great accuracy. According to statistics of open-source LLMs, their Feed-Forward Network (FFN) components typically comprise a large proportion of parameters (around $\frac{2}{3}$). This internal feature ensures that our FFN optimizations would have a better chance of achieving effective compression. Moreover, our findings are beneficial to general LLMs and are not restricted to ReLU-based models. 

This work systematically investigates the tradeoff between enforcing activation sparsity and perplexity (accuracy) on state-of-the-art LLMs. Our empirical analysis demonstrates that we can obtain around 50\% of main memory and computing reductions for critical FFN components with negligible accuracy degradation. This extra 50\% sparsity does not naturally exist in the current LLMs, which require tuning LLMs'  activation outputs by injecting zero-enforcing thresholds. To obtain the benefits of activation sparsity, we provide a guideline for the system architect for LLM prediction and prefetching. Moreover, we further verified the predictability of activation patterns in recent LLMs. The success prediction allows the system to prefetch the necessary weights while omitting the inactive ones and their successors (compress models from the memory's perspective), therefore lowering cache/memory pollution and reducing LLM execution time on resource-constraint edge devices.


\end{abstract}

\section{Introduction}

\noindent
The rapid evolution of transformed-based AI models, such as large language models (LLMs) or Large Vision Models (LVM), has significantly accelerated the recent AI expansion. State-of-the-art models typically contain several hundreds of billions or even trillion parameters (e.g., \textit{GPT-4} has \textit{1.76} trillion parameters~\cite{lubbad2023gpt} ) and deploy on HPC/cloud servers. If end users require services from these large-scale AI models, they must send requests from edge to the HPC via a network connection, and the HPC then allocates tens of high-end GPUs to execute the model inference and return results. This client-server architecture pushes all computing burdens to the centralized HPC, which not only requires stable network connections from the edge but also limits the system's scalability. Therefore,  there is a growing interest in directly deploying LLMs on edge devices~\cite{ ndhar, ullah2024role } to eliminate the above challenges. The benefits of local edge LLMs include lowering the HPC's computing and memory burdens, reducing traffic, decreasing response time, and enhancing data privacy~\cite{yin2024llm}. However, effectively deploying LLMs on edge is challenging because of their high computing and memory demands, often exceeding the capabilities of general edge devices. Big tech companies, such as \textit{Google}, \textit{Microsoft}, and \textit{Meta}, proposed smaller versions of LLMs (aka. Small Language Models (SLMs)) to deliver lightweight local AI to edge. But the abilities of SLMs can only satisfy the criteria of limited applications, and we still have great motivation to offload more powerful AI models to the edge. Moreover, the supply-demand gap between edge devices and LLMs is growing exponentially. Therefore, we must explore new compression techniques to support LLM deployment on resource-constraint devices. 

Pruning~\cite{pruning}, quantization~\cite{Yang_2019_CVPR}, and knowledge distillation ~\cite{gou2021knowledge} are three conventional methodologies to compress AI models. The compression rates theoretically may increase by utilizing one or more of these methods while maintaining appropriate accuracies. The compression rates of these approaches have theoretical up-bounds even in their ideal situations. To further attain better compression rates, we should explore a new approach that can seamlessly incorporate with existing schemes. This paper aims to investigate extra LLM compression possibilities via sparsity. AI model sparsity could be classified as \textit{weight sparsity} and \textit{activation sparsity}. Weight sparsity is more common in regular DNN models~\cite{kepner2020graphchallenge} but typically has much lower levels in state-of-the-art LLMs (Section~\ref{subsec:weight_mag_dis}). Activation sparsity~\cite{DejaVu,alizadeh2024llm} is associated with the output of FFN neurons, where a certain percentage of neurons will not be activated and no outputs pass to the subsequent layers. Therefore, from model compression's perspective, the weights of these inactive neurons are not required to be fetched into memory, which may enhance memory utilization, lower network traffic, and reduce execution latency.

The compression opportunity from activation sparsity is based on the observation that the majority of LLM's weights are from feedforward network (FFN) layers (around $\frac{2}{3}$~\cite{liu2022fl}), and not all FFN weights substantially impact the LLM output. We found that the natural activation sparsity (from default pre-trained models) of ReLU-based LLMs is normally much higher and can reach up to around \textit{95\%}. In contrast, state-of-the-art LLMs have switched to non-ReLU activations, such as \textit{SwiGLU}, demonstrating negligible natural activation sparsity. Therefore, the recent evolution of LLMs' activation functions causes new challenges for compression via activation sparsity. Similar to weight sparsity, the impact of activation values is directly associated with their magnitudes. Our investigation revealed that a significant portion of recent LLMs' activation values are extremely close to zero, and approximately 50\% of activation values in FFN layers can be safely withdrawn without significant accuracy loss. This feature allows us to predict activation patterns according to the user's input tokens. Or predict the deeper layers' patterns using the first few layers' patterns. The predicted inactive neurons, which have very small activation values, do not need to be fetched from the disk or SD card into memory. We can prefetch the active neurons to memory to lower the inference latency. Importantly, this approach does not require re-training or fine-tuning the LLMs, as the pattern prediction mechanism works on top of existing pre-trained LLMs. Our exploration also demonstrates that the activation patterns are very close (even 100\% matched) in the same LLM with similar input tokens (e.g., 70\% - 95\% similarity). This highly coincident feature indicates a huge potential to predict activation patterns and, therefore, obtain an extra ~50\% LLM compression rate. Our finding can motivate system architects to design predictors and prefetch activated weights during the execution. This optimization should significantly reduce LLM execution latency on resource-constraint edge devices because of (1) less waiting time for fetching data from disk, (2) higher memory hit rate (inactivated neurons will not be loaded into memory to avoid cache/memory pollution), (3) less computing requirements (Only compute the activated neurons).

The main contributions of this paper are outlined as follows:

\begin{itemize}

\item We explore the weight and activation sparsity of state-of-the-art LLMs. For LLMs with limited natural activation sparsity, we also examine activation magnitude distributions and learn the importance levels of FFN neurons. 

\item Based on activation magnitude distributions, we enforced activation sparsity to the state-of-the-art LLMs. To the best of our knowledge, this is the first paper to explore the enforcing activation sparsity of state-of-the-art LLMs systematically. Our empirical analysis reveals that we can secure an extra 50\% activation sparsity in FFN layers while maintaining acceptable accuracy.  

\item To convert the extra activation sparsity to compression benefits, we further investigate the predictability of activation patterns. According to our experiments, the activation patterns are highly predictable, and LLMs can be further compressed from memory's perspective. By following the activation pattern from the predictor, only the activated weights should be prefetch to memory while the inactivated ones remain in the lower-level disk. This optimization is orthogonal to existing compression techniques and may significantly reduce LLM execution time by lessening data transfer from the slow disk. 

\end{itemize}

The remaining sections of this paper are organized as follows: Section~\ref{sec:approach}  introduces our sparsity exploration methodologies. Section~\ref{sec:sparsity_finding} investigates the LLMs' natural sparsity features. Section~\ref{sec:ppl_sparsity} explores the tradeoffs between LLM enforcing activation sparsity and perplexity. Section~\ref{sec:patterns} demonstrates the similarity and predictability of activation patterns. Related work is discussed in Section~\ref{sec:related}, and finally, we conclude in Section~\ref{sec:concl}.

\section{Sparsity Exploration Approaches}
\label{sec:approach}

\subsection{LLM Selection and Evaluation Environment}
\noindent
This work targets state-of-the-art LLMs and considers the diversity of internal activation functions. Table~\ref{tab:llm_select} summarizes the pre-trained LLMs we explore, which are collected from the \textit{Hugging Face} repository.  These models were carefully chosen for their broad representation of contemporary LLM architectures, ensuring that our findings are applicable across a diverse set of current models and are not limited by the choice of architectures. For input benchmarks, we utilize the \textit{Wikitext-2}~\cite{wikitext} dataset due to its widespread adoption in LLM research, allowing easier comparison with existing studies.

\begin{table}[ht]
\centering
\caption{Selected Pre-trained LLMs}
\label{tab:llm_select}
\begin{tabular}{
  |>{\centering\arraybackslash}m{1.7cm}|
  >{\centering\arraybackslash}m{1.6cm}|
  >{\centering\arraybackslash}m{1.8cm}|
  >{\centering\arraybackslash}m{2cm}| }
\hline
\textbf{Model Name} & \textbf{Version} & \multicolumn{1}{>{\centering\arraybackslash}m{2cm}|}{\textbf{Activation Function}} & \multicolumn{1}{>{\centering\arraybackslash}m{2cm}|}{\textbf{Number of Parameters}} \\
\hline
Llama & 3 & SwiGLU & 8B \\ 
\hline
Mistral & 0.1 & SwiGLU & 7B \\
\hline
Phi & 2 & NewGELU & 2.7B \\
\hline
Phi & 3-mini-128k & SwiGLU & 3.8B \\
\hline
OPT & 1 & ReLU & 6.7B \\
\hline
\end{tabular}
\end{table}


The LLM deployment, execution, and behavior analysis are performed on our local \textit{Lambda} server, which is equipped with four \textit{NVIDIA GeForce RTX 2080 Ti} GPUs. We also developed a custom PyTorch script to log activation values and enforce sparsity during LLM inferences. This allowed us to systematically track activation features and evaluate the impact of sparsity under diversified input conditions.

\subsection{Targeting Components of LLMs for Compression}
\begin{figure}[ht]
\centering
\includegraphics[width=0.7\columnwidth]{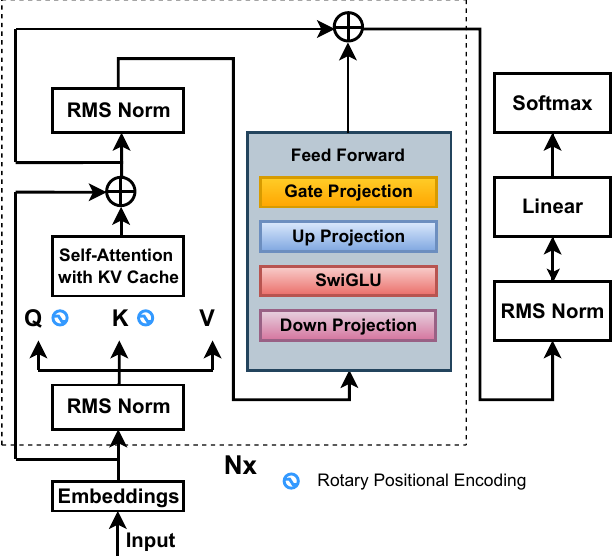}
\caption{Architecture of state-of-the-art decoder-only LLMs}
\label{fig:ffn_arch}
\end{figure}

\noindent
Popular LLMs today are based on the transformer architecture. Within this architecture,  we target the MLP (Multi-Layer Perceptron) layers, a.k.a Feed-Forward Network(FFN) layers, for model compression. As highlighted in Figure~\ref{fig:ffn_arch}, the components within FFN typically include \textit{Gate Projection}, \textit{Up Projection}, \textit{Down Projection}, and \textit{Activation\_Function}. The order of FFN layers may be varied for different LLMs. The analysis and optimizations of the FFN layers are critical for model compression and enhance the inference performance because the FFN layers contribute about $\frac{2}{3}$ of total parameters on average~\cite{liu2022fl}. Therefore, FFN layers are essentially the storage and computing bottleneck, and compressing/optimizing these components would be beneficial. 

\subsection{Enforcing Sparsity in FFN Layers}
\noindent
Our primary motivation for exploring LLMs' activation sparsity is to minimize the neuron loading from the lower-level memory hierarchy (e.r., SD card or disk) and reduce computing requirements. The potential benefits of activation sparsity are similar to those from Sparse Matrix Multiplication. The state-of-the-art LLMs with non-ReLU activation functions provide very low or no activation sparsity. Because FFN layers store most of the LLM's parameters, enforcing sparsity in FFN layers can substantially reduce the number of active neurons, leading to less computing and memory usage. However, the critical premise of enforcing activation sparsity is that the LLM accuracy should not be obviously reduced. Moreover, as another critical component of transformer-based models, attention layers are typically susceptible to being modified because they aim to capture the dependencies of all elements in a sequence. Introducing sparsity in attention layers makes it more possible to degrade model accuracy \cite{pires2023wide}. Therefore, we focus on enforcing sparsity on FFN layers to alleviate the storage and computing bottleneck. 

\begin{figure*}[ht]
  \centering
  \includegraphics[width=0.9\textwidth,height = 4.2cm]{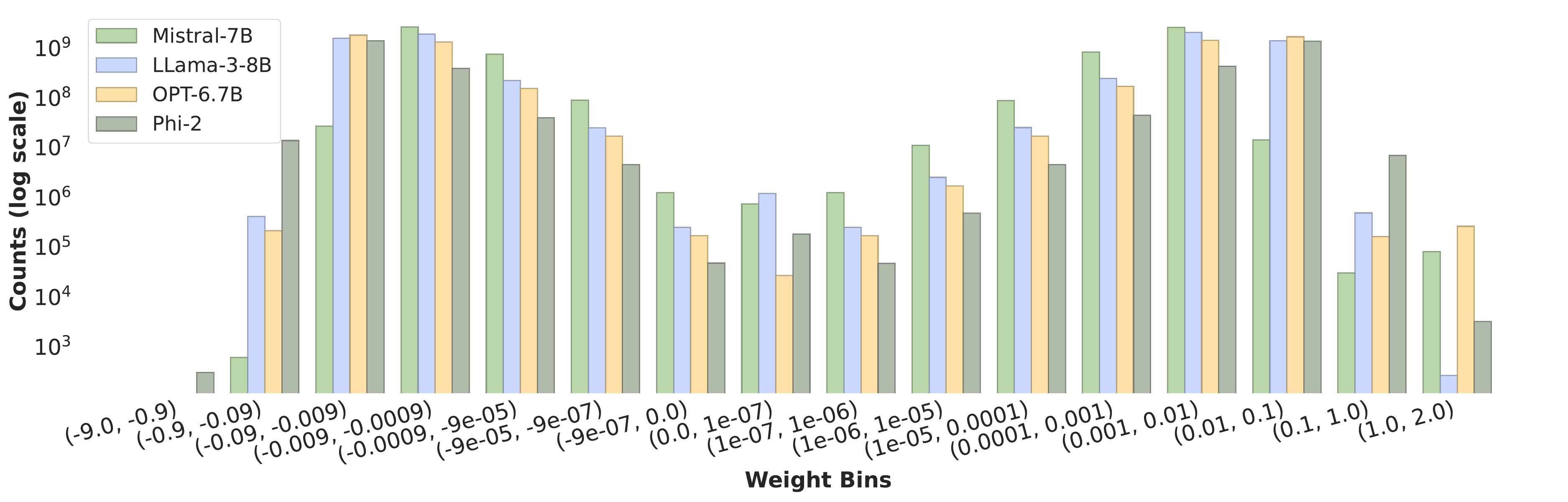}
  \caption{Weight magnitude distributions of state-of-the-art LLMs. The notation \textit{(a,b)} in the x-axis indicates a specific value range; E.g., \textit{(-9.0,-0.9)} indicates the total number of weights whose values are larger than \textit{-9.0} and smaller than \textit{-0.9}.}
  \label{fig:weight_distribute}
\end{figure*}

\subsection{Auxiliary Program for Sparsity Analysis}
\noindent
We developed an auxiliary program for the LLM sparsity analysis, which includes the following main functionalities: (1) \textit{activation behavior collection}, (2) \textit{Important Neuron Determination}, and (3) \textit{evaluations with specific thresholds}. 

\vspace{0.3em}
\subsubsection{Activation Behavior Collection}\hfill
\label{subsubsec:act_collect}

\vspace{0.3em}
We built an activation collection model consisting of the following four subcomponents. 

\vspace{0.3em}
\noindent\underline{\textit{Input Sequence}}: Let \( \mathbf{X} = \{ \mathbf{x}_1, \mathbf{x}_2, \ldots, \mathbf{x}_N \} \) represent the LLM input tokens where \( N \) is the number of tokens.

\vspace{0.3em}
\noindent\underline{\textit{Forward Pass Inference}}: For each input token \( \mathbf{x}_i \), perform a forward pass through the model to obtain the activation values \( \mathbf{A}_{i,l} \) for each layer \( l \):
\begin{equation}
\mathbf{A}_{i,l} = f_l(\mathbf{x}_i)
\end{equation}
where \( f_l \) is the activation function of layer \( l \) and $i \in [1,N]$.

\vspace{0.3em}
\noindent\underline{\textit{Register Hooks}}: For each layer \( l \), register hooks to capture the activation values during the inference. We define \( \mathbf{A}_l \) as the activations for layer \( l \) across all input tokens:
\begin{equation}
\mathbf{A}_l = \{ \mathbf{A}_{i,l} \mid i \in \{1, 2, \ldots, N\} \}
\end{equation}
\noindent\underline{\textit{Activation Storage}}:
Store the collected activation values \( \mathbf{A}_l \) in an HDF5 file for next-step processing.

\vspace{0.3em}
\subsubsection{Important Neuron Determination}\hfill

\vspace{0.3em}
Following the \textit{Activation Behavior Collection} step discussed in Section~\ref{subsubsec:act_collect}, we must define the approach to select a specific percentage of neurons with smaller output magnitudes that are less important. The magnitude is directly associated with the importance of the neuron. Therefore, enforcing sparsity to the smaller magnitude neurons should have minimum difference compared to original output. The \textit{Important Neuron Determination} contains the following substeps.

\vspace{0.3em}
\noindent\underline{\textit{Flatten Activation Matrix to Single-Dimension Vector}}: \hfill 
For each layer \( l \), flatten the activation matrix values \( \mathbf{A}_l \) into a single dimension vector \( \mathbf{A}_l^\text{flat} \).

\begin{equation}
\mathbf{A}_l^\text{flat} = \text{flatten}(\mathbf{A}_l)
\end{equation}
\noindent\underline{\textit{Converting to Absolute Values}}:
Compute the absolute values of the flattened activation vector. The importance of activation values are directly associated with magnitudes instead of signs:
\begin{equation}
\mathbf{A}_l^\text{abs} = \left| \mathbf{A}_l^\text{flat} \right|
\end{equation}

\noindent\underline{\textit{Sorting}}:
Sort the vector with absolute values in ascending order to facilitate the following percentile computation:
\begin{equation}
\mathbf{A}_l^\text{sorted} = \text{sort}(\mathbf{A}_l^\text{abs})
\end{equation}

\noindent\underline{\textit{Percentile Calculation}}:
For a given sparsity level \( \alpha \) (e.g., 30\%), calculate the threshold \( T_{l,\alpha} \) as the lowest \( \alpha \)-th percentile of \( \mathbf{A}_l^\text{abs} \). \( P_{\alpha} \) denotes as the percentile function at sparsity \( \alpha \).
\begin{equation}
T_{l,\alpha} = P_{\alpha}(\mathbf{A}_l^\text{sorted})
\end{equation}

\subsubsection{Evaluations with Specific Sparsity Thresholds}\hfill

\vspace{0.3em}
\noindent
After applying a pre-defined activation threshold, we obtain a new LLM with a different sparsity. Specifically, we implemented a mechanism to deactivate neurons in \textit{gate projection}, \textit{up projection}, and \textit{down projection} components whose values fell below the threshold. Activation values are set to zero if they are less than \( T_{l,\alpha} \) in magnitude:

\begin{equation}
\mathbf{A}_{i,l}^\text{thresholded} = \begin{cases} 
0 & \text{if } \left| \mathbf{A}_{i,l} \right| < T_{l,\alpha} \\
\mathbf{A}_{i,l} & \text{otherwise}
\end{cases}
\end{equation}

\section{Sparsity Findings of State-of-the-art LLMs}
\label{sec:sparsity_finding}

\subsection{Weight Magnitude Distributions}
\label{subsec:weight_mag_dis}

\noindent
Before systematically studying the LLMs' sparsity, exploring their weight magnitude distributions is a prerequisite to understanding internal features. Figure~\ref{fig:weight_distribute} illustrates the weight distributions of four recent LLMs: \textit{Mistral-7B}, \textit{LLama-3-8B}, \textit{OPT-6.7B}, and \textit{Phi-2-2.7B} 
The x-axis represents sequential weight bins; each bin shows the weight count in a specific range. The bin ranges are denoted in value pairs, indicating the start and end of each range. Bin ranges in Figure~\ref{fig:weight_distribute} span from \textit{(-9.0, -0.9)} to \textit{(1.0, 2.0)}, all weights of our evaluated models fall into these ranges. The y-axis displays the weight counts on a logarithmic scale. 

\vspace{2pt}
\noindent \textbf{Observation}: Upon analyzing the weight magnitude distributions of LLMs in Figure~\ref{fig:weight_distribute}, we can obtain an essential guideline for LLM pruning or enforcing activation sparsity. The distribution of weights across different bins reveals that a large proportion of the weights fall within the small magnitude ranges, and a very small number of weights are absolute zeros. This trend is observed in all four evaluated LLMs, including \textit{Mistral-7B}, \textit{LLama-3-8B}, \textit{OPT-6.7B}, and \textit{Phi-2-2.7B}. In other words, the state-of-the-art LLMs do not directly provide significant weight sparsity, even though the magnitudes of a considerable percentage of weights are very close to zero.

\begin{figure}[t]
\centering
\includegraphics[width=1\columnwidth,height = 3.8cm]{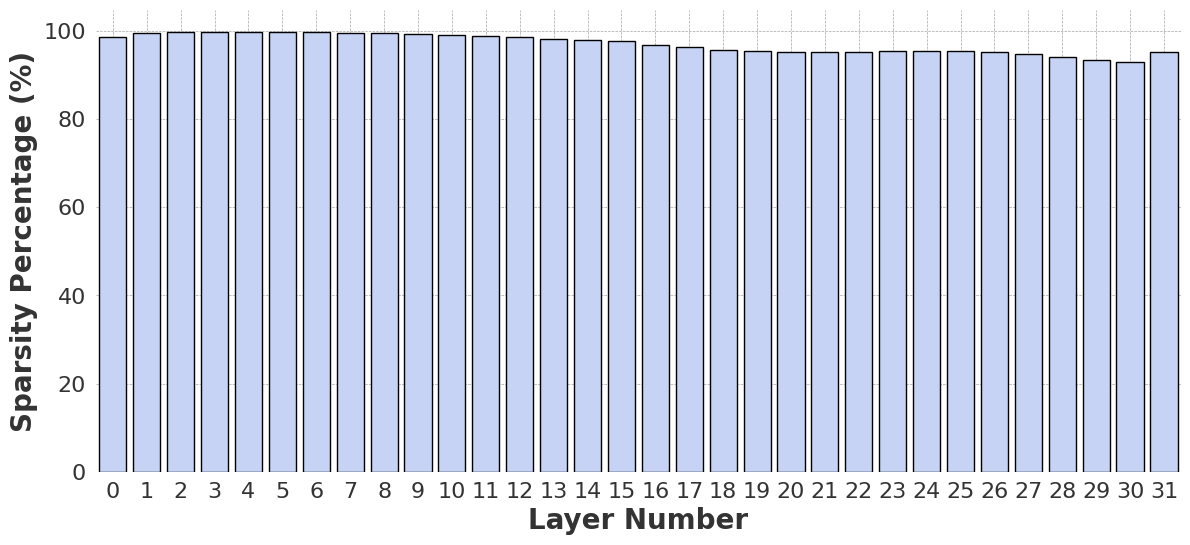}
\caption{Natural activation sparsity of all layers in \textit{OPT-6.7B}}
\label{fig:OPT_Activation_Sparsity}
\end{figure}

\vspace{2pt}
\noindent \textbf{\underline{Insight}}: Unlike some conventional DNNs~\cite{kepner2020graphchallenge}, we do not obverse substantial weight sparsity in four recent LLMs. Therefore, to compress LLM via the sparsity feature, we must explore another sparsity type: activation sparsity. 

\subsection{Natural Activation Sparsity Analysis}
\label{subsec:act_sparsisty_analysis}
\noindent
Activation sparsity could be another essential feature and opportunity to compress the LLMs, especially when deploying these large-scale AI models in resource-constrained environments. The sparsity feature will not only provide opportunities to reduce memory storage but also lower the computing resource requirements. This improvement eventually optimizes the edge LLM execution latency, which provides quick input responses and better user experiences. We show the results of \textit{OPT-6.7B} and \textit{Phi-2-2.7B} in this section, as they are the only two models that exhibit natural sparsity in their Feed-Forward Network (FFN) activation values in our study. We do not observe any natural sparsity in other LLMs (\textit{Mistral-7B}, \textit{LLama-3-8B} and \textit{Phi-3-3.8B}) we evaluated based on WikiText. The term '\textit{natural activation sparsity}' indicates the activation sparsity directly comes from the pre-trained LLM instead of from enforcing sparsity.

\vspace{2pt}
\noindent \textbf{Observation 1 (\textit{OPT-6.7B})}: The activation sparsity histogram of FFN layers for the \textit{OPT-6.7B} is demonstrated in Figure~\ref{fig:OPT_Activation_Sparsity}. The sparsity percentages from \textit{Layer 0} to \textit{Layer 31} are all high, even though a slight sparsity degradation is observed in deeper layers. \textit{Layer 30} has \textit{92.83}\% natural sparsity, which is the minimal sparsity among all layers. According to Table~\ref{tab:llm_select}, the activation function of  \textit{OPT-6.7B} is ReLU. The ReLU activation function's outputs are set to zero if the neuron inputs are less than zero. This attribute of the ReLU function is the main reason for obtaining high natural sparsity. Apple's recent work~\cite{alizadeh2024llm} for LLM compression is based on a similar observation and targeting to models with the ReLU activation.

\begin{figure}[t]
\centering
\includegraphics[width=1\columnwidth, height = 3.8cm]{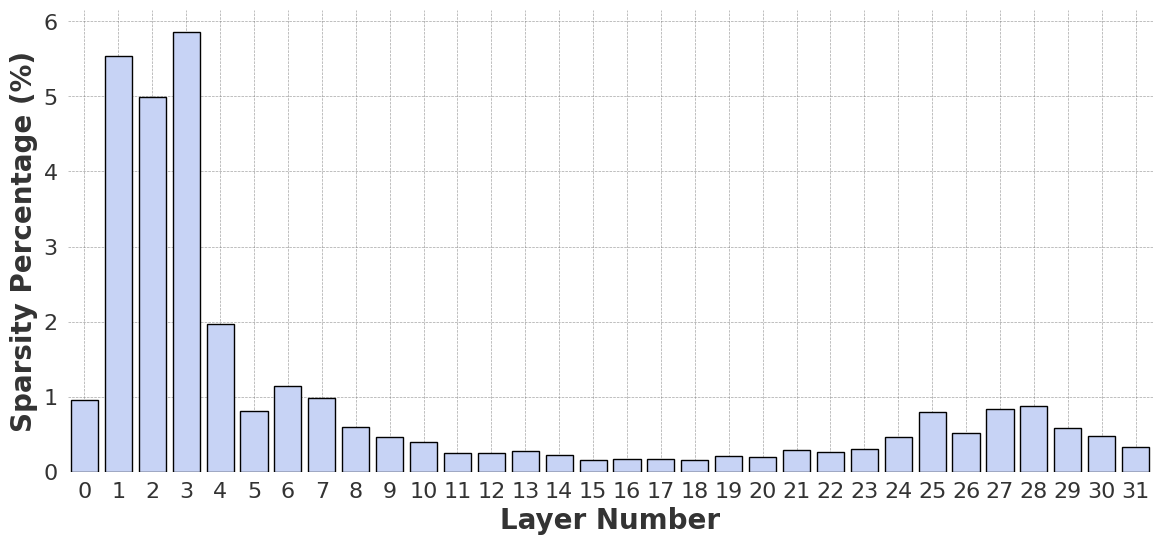}
\caption{Natural activation sparsity of all layers in \textit{Phi-2-2.7B}}
\label{fig:Phi-2_Activation_Sparsity}
\end{figure}

\vspace{2pt}
\noindent \textbf{Observation 2 (\textit{Phi-2-2.7B})}: Figure ~\ref{fig:Phi-2_Activation_Sparsity} illustrates the activation sparsity of FFN layers for the \textit{Phi-2-2.7B} model. Even though a certain degree of sparsity exists, the maximum sparsity level (\textit{Layer 3}) is still less than 6\%. The sparsity percentages of the majority of layers are less than 1\%. Compared to the \textit{OPT-6.7B},  \textit{Phi-2-2.7B} shows much lower activation sparsity. The main reason for this discrepancy is their internal activation functions, where \textit{OPT-6.7B} utilizes ReLU while \textit{Phi-2-2.7B} uses NewGELU. By comparing their equation curves, ReLU has a significantly higher possibility of obtaining a '0' activation output.

\vspace{2pt}
\noindent \textbf{\underline{Insight}}: The benefits of natural activation sparsity only exist in ReLU-based LLMs (e.g., \textit{OPT-6.7B}). Even though we can also observe natural activation sparsity in NewGELU-based LLMs (e.g., \textit{OPT-6.7B}), the sparsity level is too low to make effective model compression. The state-of-the-art LLMs mainly use SwiGLU function. Therefore, we should explore other new and general methods to obtain extra activation sparsity for LLM compression.

\subsection{Activation Magnitude Distributions}
\label{subsec:act_mag_dist}
\noindent
From the discussion in Section~\ref{subsec:act_sparsisty_analysis}, most state-of-the-art LLMs do not exhibit inherent natural activation sparsity due to the old function ReLU or NewGELU has been replaced. Significant portions of activation values in recent LLMs do not reach absolute zero. Therefore, this section examines activation magnitude distributions for other potential sparsity opportunities. For example, according to the results of activation magnitude distributions, can we adjust the sparsity level by modifying a certain percentage of small magnitude values to zero and increasing the activation sparsity (Section~\ref{sec:ppl_sparsity})? Exploring activation magnitude distributions is a crucial prerequisite step to answer this question. So, the distributions of (\textit{Phi-3-3.8B},  \textit{LLaMA-3-8B} and \textit{Mistral-7B}) are shown and analyzed below. These three recent LLMs are non-ReLU-based and do not exhibit natural sparsity.

\begin{figure}[ht]
\centering
\includegraphics[width=1\columnwidth]{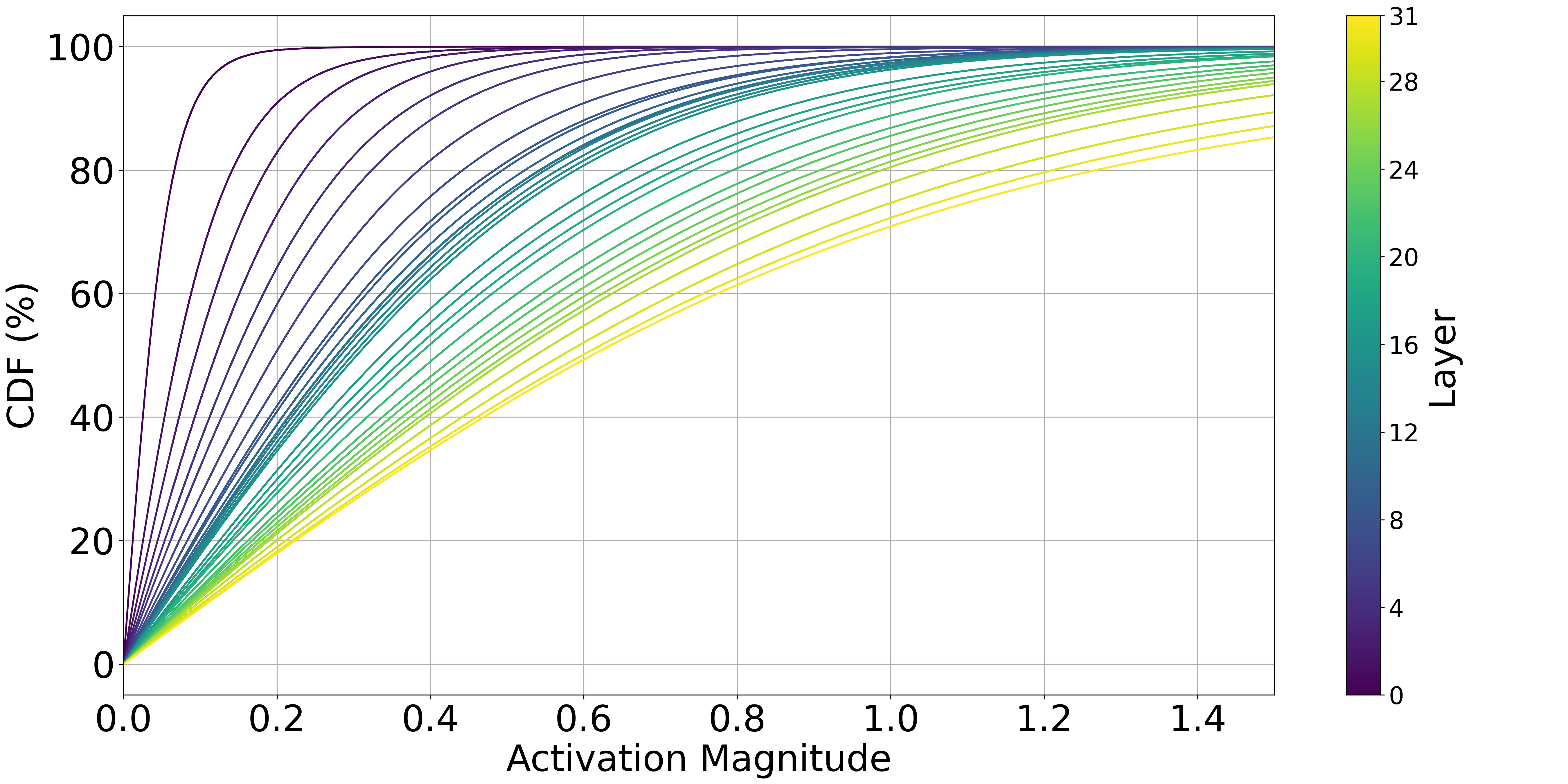}
\caption{\centering Activation magnitude distribution of Phi-3-3.8B for all 32 layers (Gate-Up Projection)}
\label{fig:Phi-3_cdf}
\end{figure}

\vspace{2pt}
\noindent \textbf{Observation 1 (\textit{Phi-3-3.8B}):} Figure~\ref{fig:Phi-3_cdf} displays the activation cumulative distribution function (CDF) of the \textit{Gate-Up Projection} activation components for the \textit{Phi-3-3.8B} across all its 32 layers. A significant concentration of activation values falls below \textit{0.05}, particularly in the first layer (\textit{Layer 0}), where approximately 60\% of activation values are under \textit{0.05}. In the subsequent three layers (\textit{Layer 1 - 3}), about 70\% of activation values range from 0.0 to 0.2, indicating a slight dispersion. As moving beyond these first several layers, activation values are more evenly spread over relatively wider ranges. The CDF further suggests that setting a cutoff threshold at \textit{0.6} for \textit{Phi-3-3.8B} could result in around 50\% activation reduction, providing the potential for extra activation sparsity.

\begin{figure}[ht]
\centering 
\includegraphics[width=1\columnwidth]{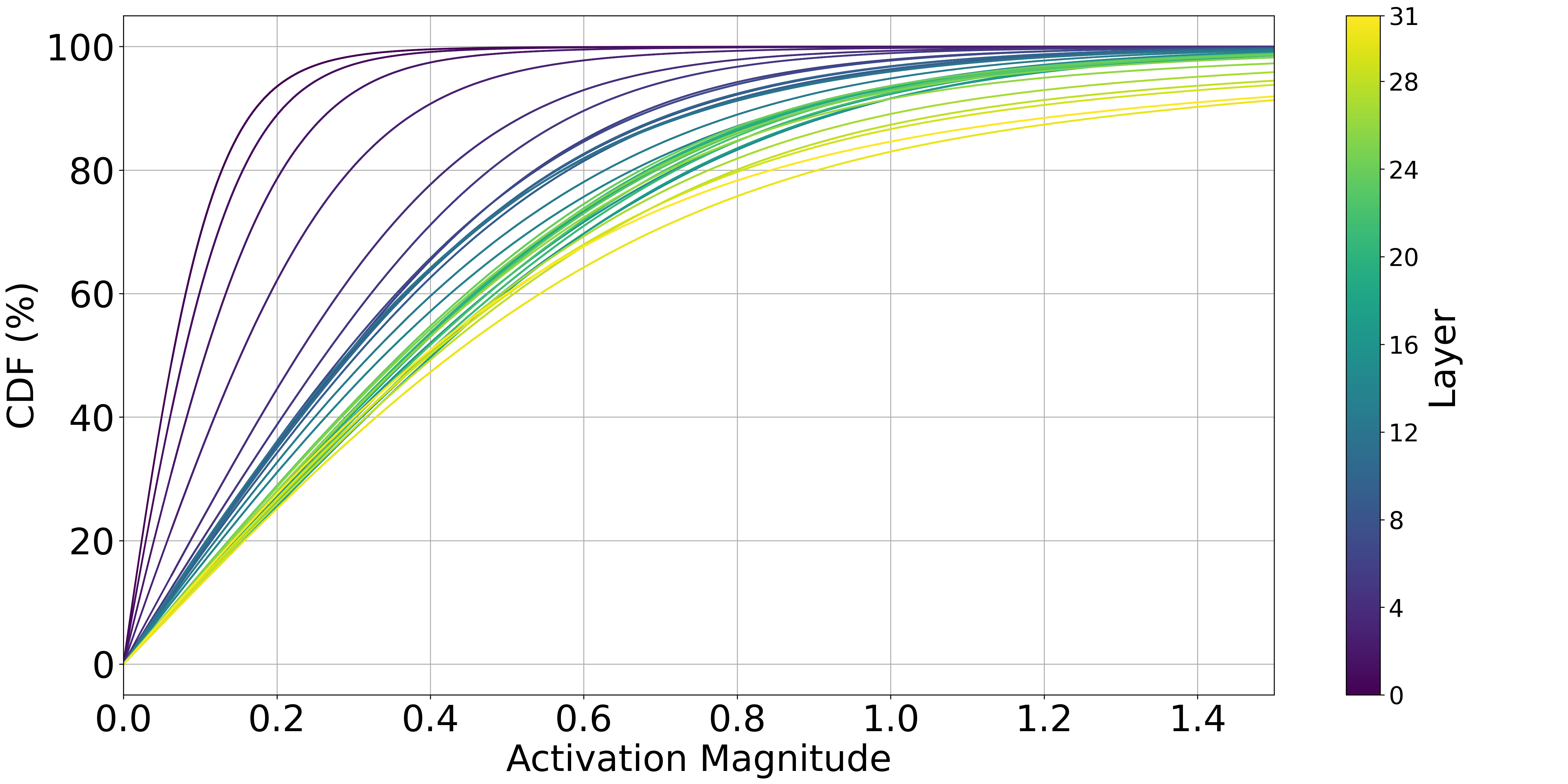}
\caption{\centering Activation magnitude distribution of LLaMA-3-8B for all 32 layers (Gate Projection)}
\label{fig:LLama-3_cdf}
\end{figure}

\vspace{2pt}
\noindent \textbf{Observation 2 (\textit{LLaMA-3-8B}):} Figure~\ref{fig:LLama-3_cdf} presents the activation CDF of \textit{Gate Projection} component of \textit{LLaMA-3-8B} in all its 32 layers. For \textit{Layer 0}, 65\% of activation values fall below \textit{0.1}. For the subsequent three layers (\textit{Layer 1 - 3}), more than 60\% of activation values are below \textit{0.2}. When examining the remainder of layers, the graph suggests that setting a threshold at \textit{0.5} would allow us to discard 50\% of neurons in FFN, implying a potential for 50\% activation sparsity. 

\begin{figure}[ht]
\centering
\includegraphics[width=1\columnwidth]{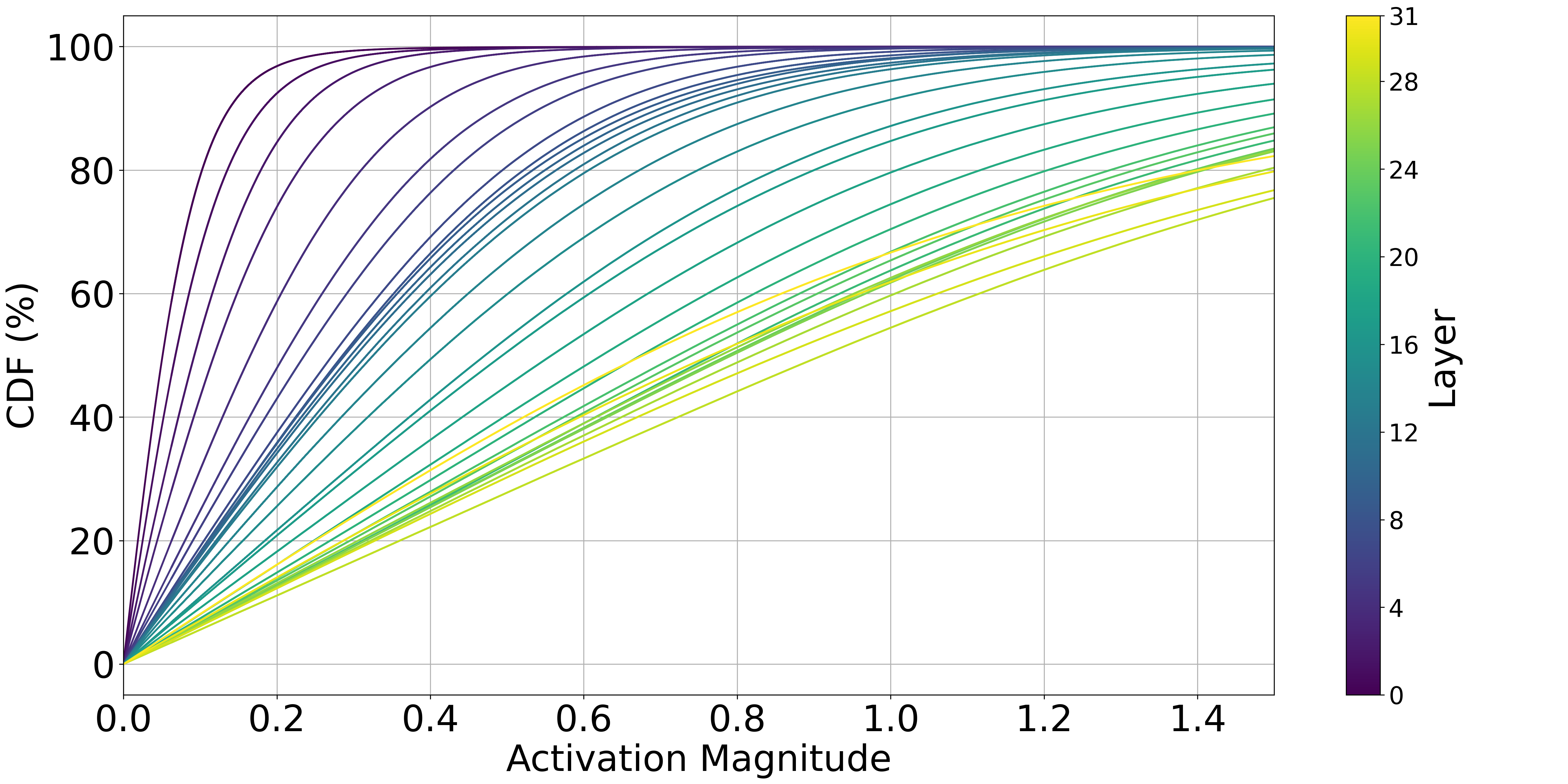}
\caption{\centering Activation magnitude distribution of Mistral-7B for all 32 layers (Gate Projection)}
\label{fig:Mistral-7B_CDF}
\end{figure}

\vspace{2pt}
\noindent \textbf{Observation 3 (\textit{Mistral-7B}):} Figure~\ref{fig:Mistral-7B_CDF} reveals the activation CDF of the \textit{Gate Projection} component of \textit{Mistral-7B} in all its 32 layers. In \textit{Layer 0}, approximately 80\% of the activation values are below \textit{0.1}. From \textit{Layer 1} to \textit{Layer 3}, around 80\% of the activation values lie below \textit{0.25}. For the deeper layers, the CDF indicates that applying a cutoff threshold at \textit{1.1} would potentially omit 60\% of the FFN neurons to obtain extra sparsity. The trends of activation magnitude distributions for the three evaluated LLMs are similar. The first few layers consist of smaller magnitude activation outputs. For the deeper layers, the magnitude distribution gradually becomes more even, but they are still limited within relatively small ranges.

\vspace{2pt}
\noindent \textbf{\underline{Insight}}: The magnitudes of majority activation values fall into very small data ranges. This observation allows us to set small thresholds to omit fewer contribution weights and easily obtain high sparsity levels.   

\section{Tradeoffs between Activation Sparsity and Perplexity}
\label{sec:ppl_sparsity}
\begin{figure}[t]
\centering
\includegraphics[width=0.9\columnwidth]{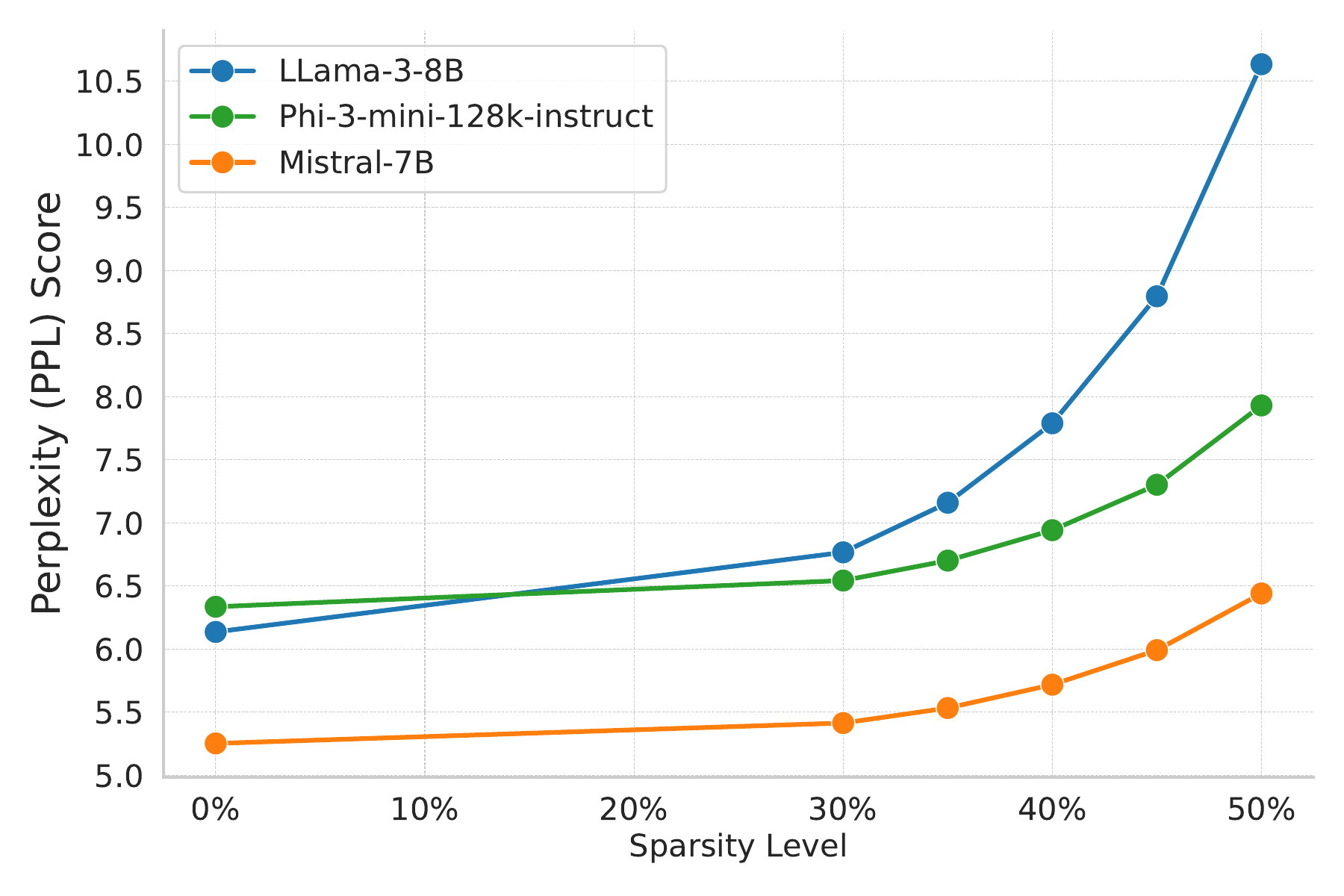}
\caption{\centering Tradeoffs between Activation Sparsity and Perplexity (PPL) Scores}
\label{fig:ppl}
\end{figure}

\noindent
Section~\ref{subsec:act_mag_dist} concludes that we can obtain a high sparsity level by enforcing relatively small threadholds. However, the new question for enforcing aviation sparsity is how much accuracy degradation will be observed if we get this extra sparsity?  This section investigates how much sparsity we can tune while maintaining an acceptable accuracy degradation. The sparsity tuning allows us to make more effective compression and support more powerful LLMs on edge systems. The LLM accuracies are evaluated by the Perplexity (PPL) scores. Similarly to Section~\ref{subsec:act_mag_dist}, we shows the results of \textit{Phi-3-3.8B},  \textit{LLaMA-3-8B} and \textit{Mistral-7B} for consistency. The \textit{x-axis} represents the sparsity level, ranging from 0\% to 50\%, indicating the percentage of activations enforced to become zero. The \textit{y-axis} shows the Perplexity (PPL) score, where lower scores indicate better accuracy when executing the \textit{Wikitext} benchmark.

\vspace{2pt}
\noindent \textbf{Observation:} Figure~\ref{fig:ppl} illustrates tradeoffs between activation sparsity and perplexity (PPL) scores. As the sparsity level increases, the PPL generally rises for all models, indicating that increasing LLM activation sparsity reduces inference accuracy. At 30\% sparsity level, all models almost keep the same PPL with 0\% sparsity. In other words, all three models can obtain 30\% sparsity with only negligible accuracy loss. \textit{LLama-3-8B} shows the steepest increase in PPL, indicating the most significant accuracy drop in performance as sparsity increases. However, even for 50\% sparsity, the maximum PPL from \textit{LLama-3-8B} is still less than \textit{11}, which is considered acceptable in many application scenarios. In contrast, \textit{Mistral-7B}'s accuracy is most stable when changing activation sparsity.

\vspace{2pt}
\noindent \textbf{\underline{Insight}}: We can obtain an extra 50\% activation sparsity for the state-of-the-art LLMs by enforcing the threshold setting while maintaining acceptable PPL/accuracy.

\section{Predictability Analysis for LLM Activation Patterns}
\label{sec:patterns}
\noindent
According to our experimental data and analysis in Section~\ref{sec:ppl_sparsity}, we conclude that we can enforce around 50\% activation sparsity to state-of-the-art LLMs while maintaining almost the same accuracy/perplexity. A straightforward way to utilize activation sparsity to compress LLMs is via activation pattern prediction. We can predict the future activation pattern and only fetch the possible active weights from the disk or SD card to the main memory. A reasonably predicted success rate can provide benefits, including lower LLM response latency, fewer computing resource requirements, and better main memory utilization. Dhar et al.~\cite{ndhar} indicate that memory is the essential bottleneck for LLM edge deployment, significantly increasing the LLM execution latency on resource-constraint devices. Success predictions have great potential to alleviate memory bottlenecks. The incorrect prediction has an extra latency penalty because the system must fetch the mispredicted weights from the disk to the main memory. However, the misprediction in this problem is not a correctness issue. This section will systematically explore LLM activation patterns and analyze their predictability. We aim to provide a guideline for system architects to design an effective prefetcher to assist LLM deployment on edge. 


\begin{table}[ht]
\centering
\caption{Activation Pattern Match Rates for Similar Inputs}
\label{tab:similarity}
\resizebox{\columnwidth}{!}{
\small 
\begin{tabular}{
  |>{\centering\arraybackslash}m{1cm}|
  >{\centering\arraybackslash}m{0.6cm}|
  >{\centering\arraybackslash}m{0.6cm}|
  >{\centering\arraybackslash}m{0.6cm}|
  >{\centering\arraybackslash}m{0.6cm}|
  >{\centering\arraybackslash}m{0.6cm}|
  >{\centering\arraybackslash}m{0.6cm}| }
\hline
\multirow{2}{*}{\textbf{Sample}} & \multicolumn{6}{c|}{\textbf{Similarity Percentages of Input Variants}} \\
\cline{2-7}
 & \textbf{95\%} & \textbf{90\%} & \textbf{85\%} & \textbf{80\%} & \textbf{75\%} & \textbf{70\%} \\
\hline
\textbf{1} & 100\% & 100\% & 100\% & 100\% & 100\% & 100\% \\
\hline
\textbf{2} & 57\% & 57\% & 57\% & 57\% & 57\% & 57\% \\
\hline
\textbf{3} & 100\% & 100\% & 100\% & 100\% & 100\% & 100\% \\
\hline
\textbf{4} & 100\% & 53\% & 53\% & 53\% & 53\% & 53\% \\
\hline
\textbf{5} & 100\% & 100\% & 100\% & 100\% & 100\% & 100\% \\
\hline
\textbf{6} & 100\% & 100\% & 100\% & 100\% & 100\% & 100\% \\
\hline
\textbf{7} & 100\% & 100\% & 100\% & 100\% & 100\% & 100\% \\
\hline
\textbf{8} & 57\% & 57\% & 57\% & 57\% & 57\% & 57\% \\
\hline
\textbf{9} & 100\% & 100\% & 100\% & 100\% & 100\% & 100\% \\
\hline
\textbf{10} & 100\% & 100\% & 100\% & 100\% & 100\% & 100\% \\
\hline
\textbf{11} & 100\% & 100\% & 100\% & 100\% & 100\% & 100\% \\
\hline
\textbf{12} & 100\% & 100\% & 100\% & 100\% & 100\% & 100\% \\
\hline
\end{tabular}}
\end{table}


\begin{figure*}[ht]
     \centering
     \begin{subfigure}[b]{0.24\textwidth}
         \centering
         \includegraphics[width=\textwidth]{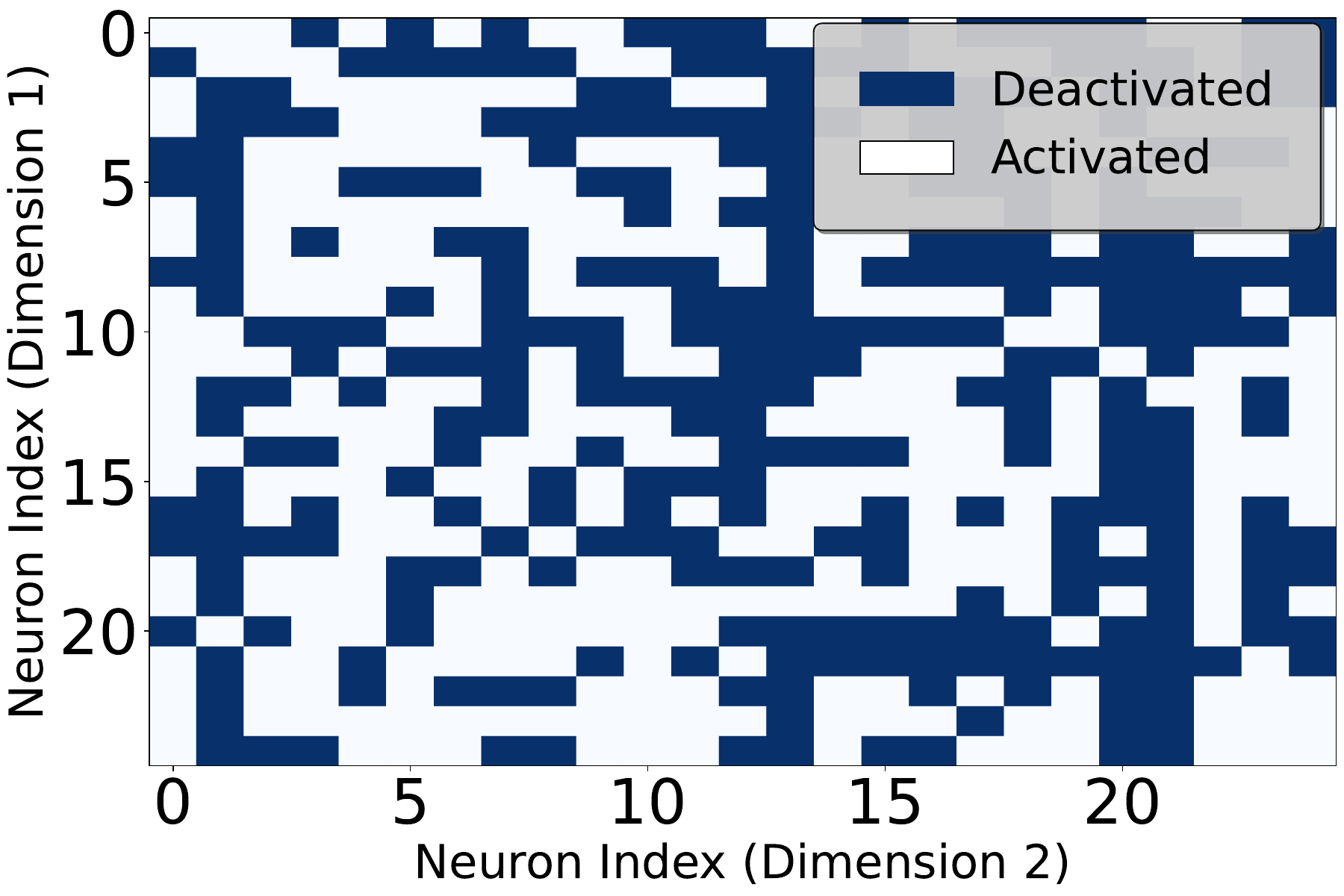}
         \caption{\centering Layer 1}
         \label{fig:Sample1_100_layer1}
     \end{subfigure}
     \hfill
     \begin{subfigure}[b]{0.24\textwidth}
         \centering
         \includegraphics[width=\textwidth]{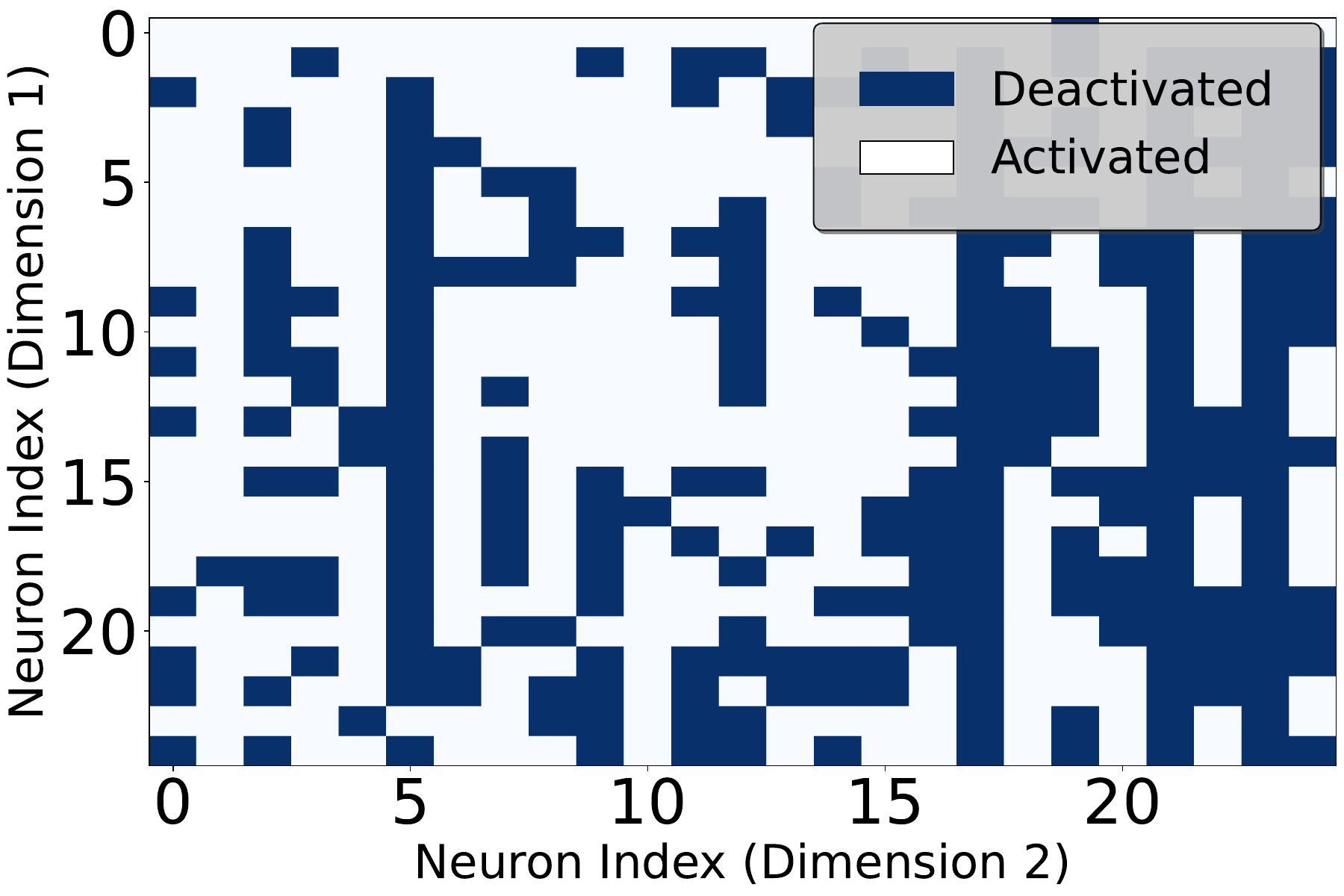}
         \caption{\centering Layer 10}
         \label{fig:Sample1_100_layer10}
     \end{subfigure}
     \hfill
     \begin{subfigure}[b]{0.24\textwidth}
         \centering
         \includegraphics[width=\textwidth]{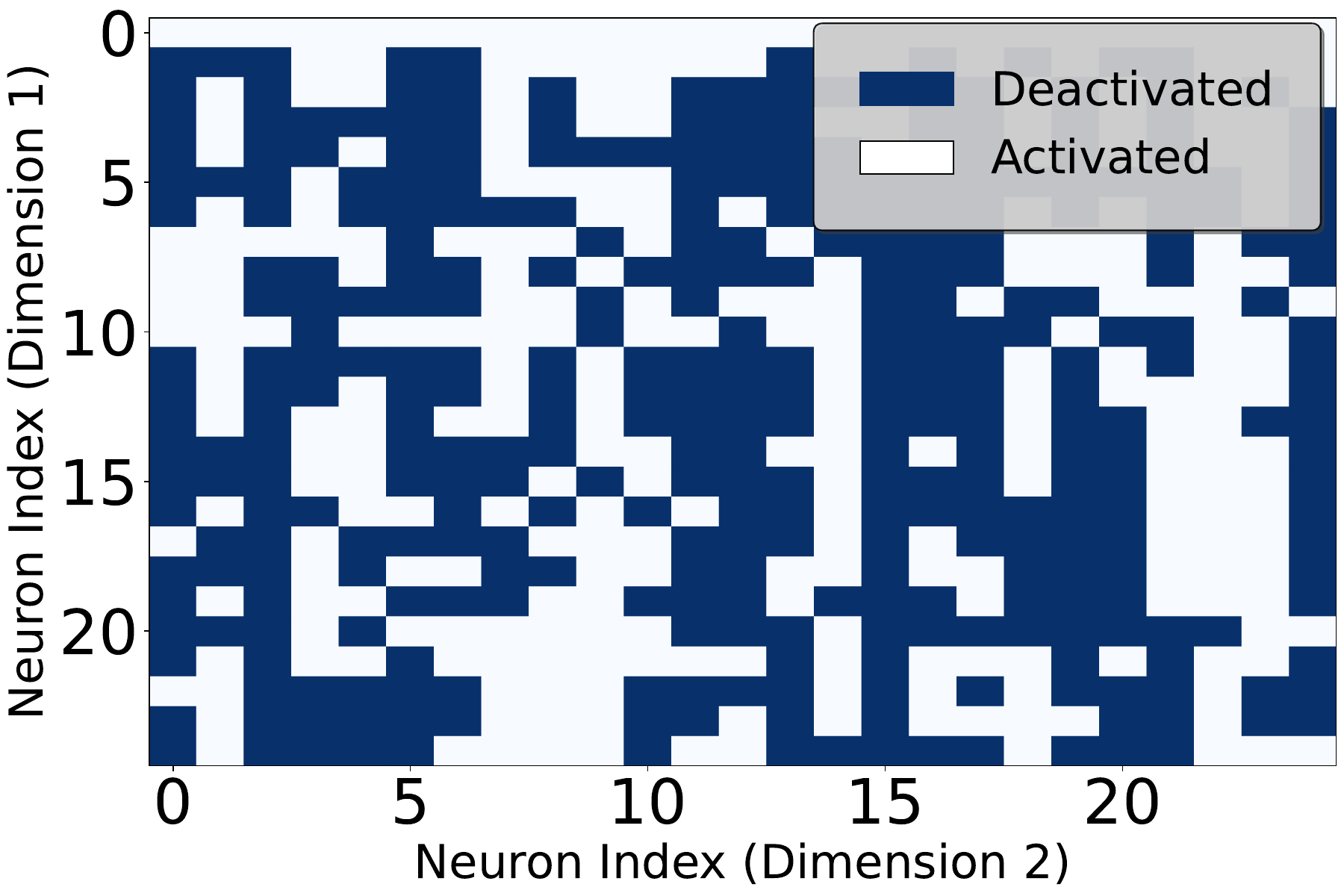}
         \caption{\centering Layer 20}
         \label{fig:Sample1_100_layer20}
     \end{subfigure}
     \hfill
     \begin{subfigure}[b]{0.24\textwidth}
         \centering
         \includegraphics[width=\textwidth]{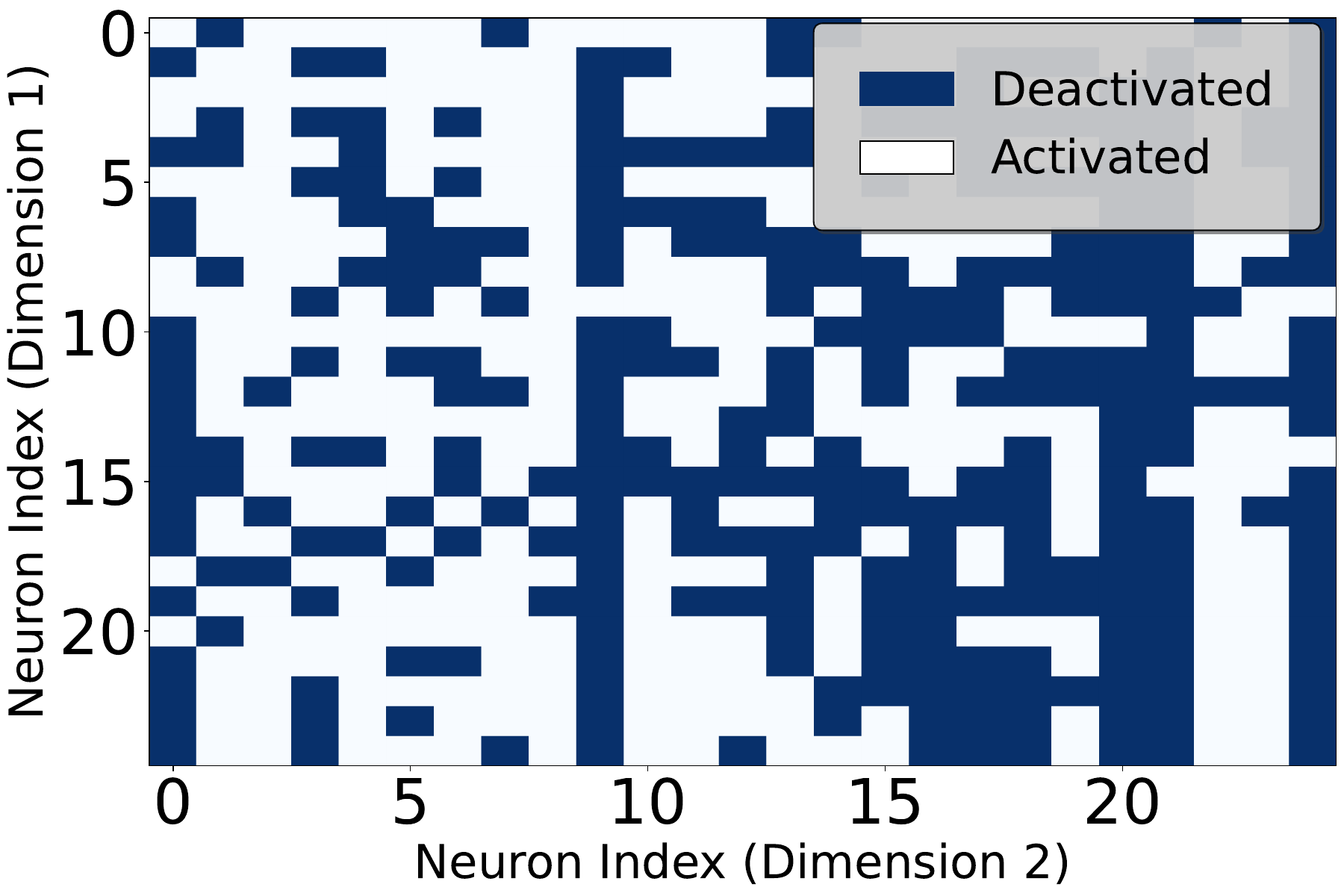}
         \caption{\centering Layer 32}
         \label{fig:Sample1_100_layer32}
     \end{subfigure}
        \vspace{-0.3em}
        \caption{Activation heatmap pattern of LLaMa-3-8b with default Sample 1 input}
        \label{fig:Sample_1_default}
        \vspace{-0.3em}
\end{figure*}

\begin{figure*}[ht]
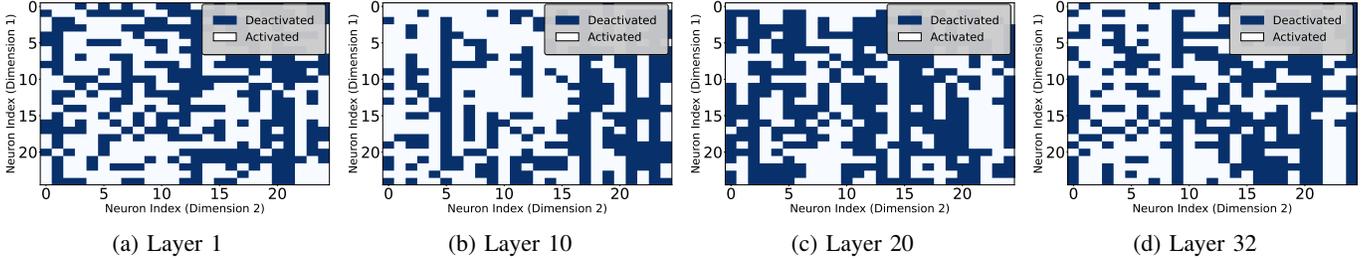

     \centering
     \begin{subfigure}[b]{0.24\textwidth}
         \centering
         \includegraphics[width=\textwidth]{figs/fig_source/sample1/activation_heatmap_para8_layer_1_similarity70.pdf}
         \caption{\centering Layer 1}
         \label{fig:Sample1_70_layer1}
     \end{subfigure}
     \hfill
     \begin{subfigure}[b]{0.24\textwidth}
         \centering
         \includegraphics[width=\textwidth]{figs/fig_source/sample1/activation_heatmap_para8_layer_10_similarityOrigin.pdf}
         \caption{\centering Layer 10}
         \label{fig:Sample1_100_layer10}
     \end{subfigure}
     \hfill
     \begin{subfigure}[b]{0.24\textwidth}
         \centering
         \includegraphics[width=\textwidth]{figs/fig_source/sample1/activation_heatmap_para8_layer_20_similarity70.pdf}
         \caption{\centering Layer 20}
         \label{fig:Sample1_70_layer20}
     \end{subfigure}
     \hfill
     \begin{subfigure}[b]{0.24\textwidth}
         \centering
         \includegraphics[width=\textwidth]{figs/fig_source/sample1/activation_heatmap_para8_layer_32_similarity70.pdf}
         \caption{\centering Layer 32}
         \label{fig:Sample1_70_layer32}
     \end{subfigure}
        \vspace{-0.3em}
        \caption{\centering Activation heatmap pattern of LLaMa-3-8b with 70\% similarity Sample 1 input; Comparing with default Sample 1's patterns in Figure~\ref{fig:Sample_1_default}, matching rates are 100\% for all evaluated layers}
        \label{fig:Sample_1_70}
        \vspace{-0.3em}
\end{figure*}

\begin{figure*}[ht]
     \centering
     \begin{subfigure}[b]{0.24\textwidth}
         \centering
         \includegraphics[width=\textwidth]{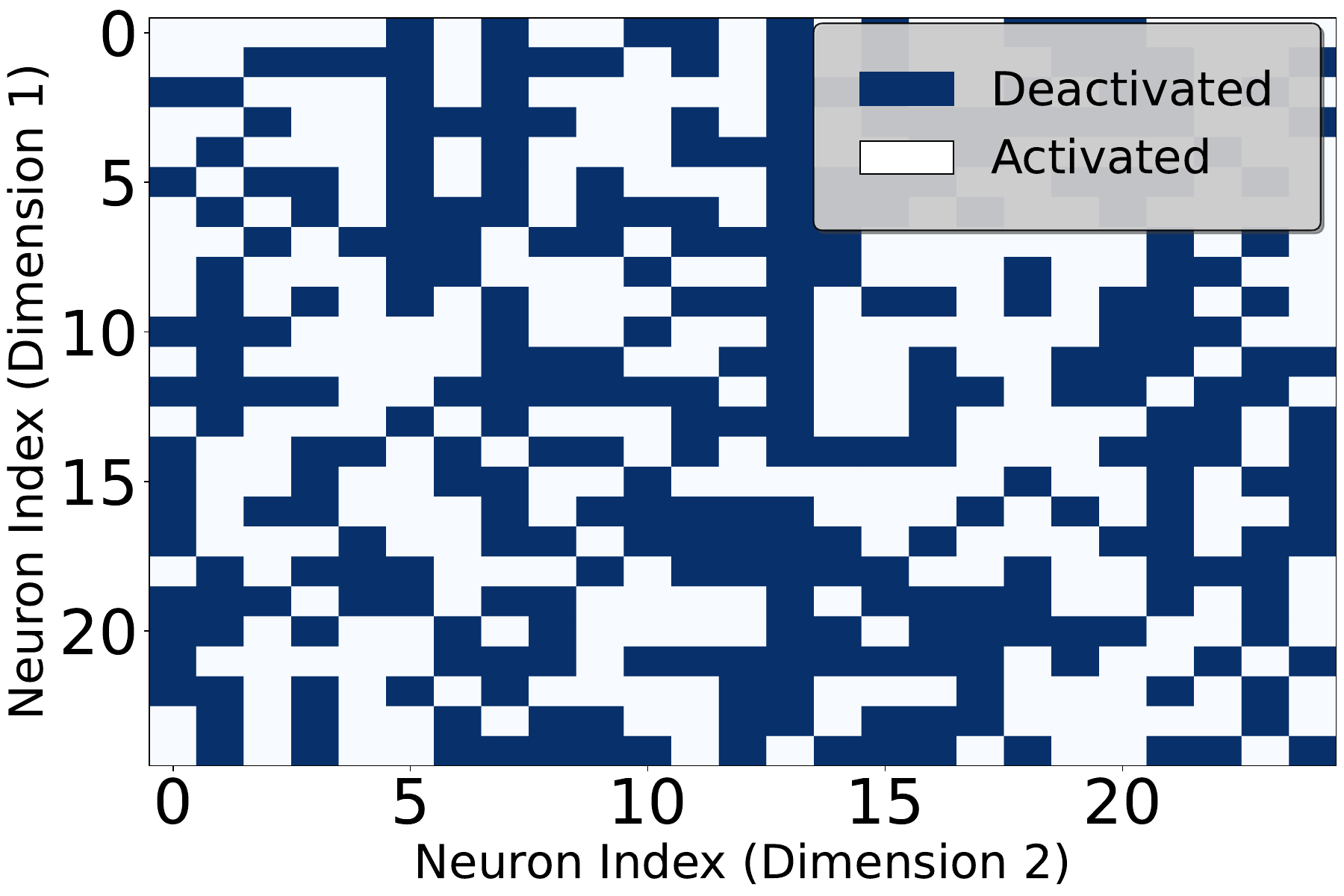}
         \caption{\centering Layer 1}
         \label{fig:Sample8_100_layer1}
     \end{subfigure}
     \hfill
     \begin{subfigure}[b]{0.24\textwidth}
         \centering
         \includegraphics[width=\textwidth]{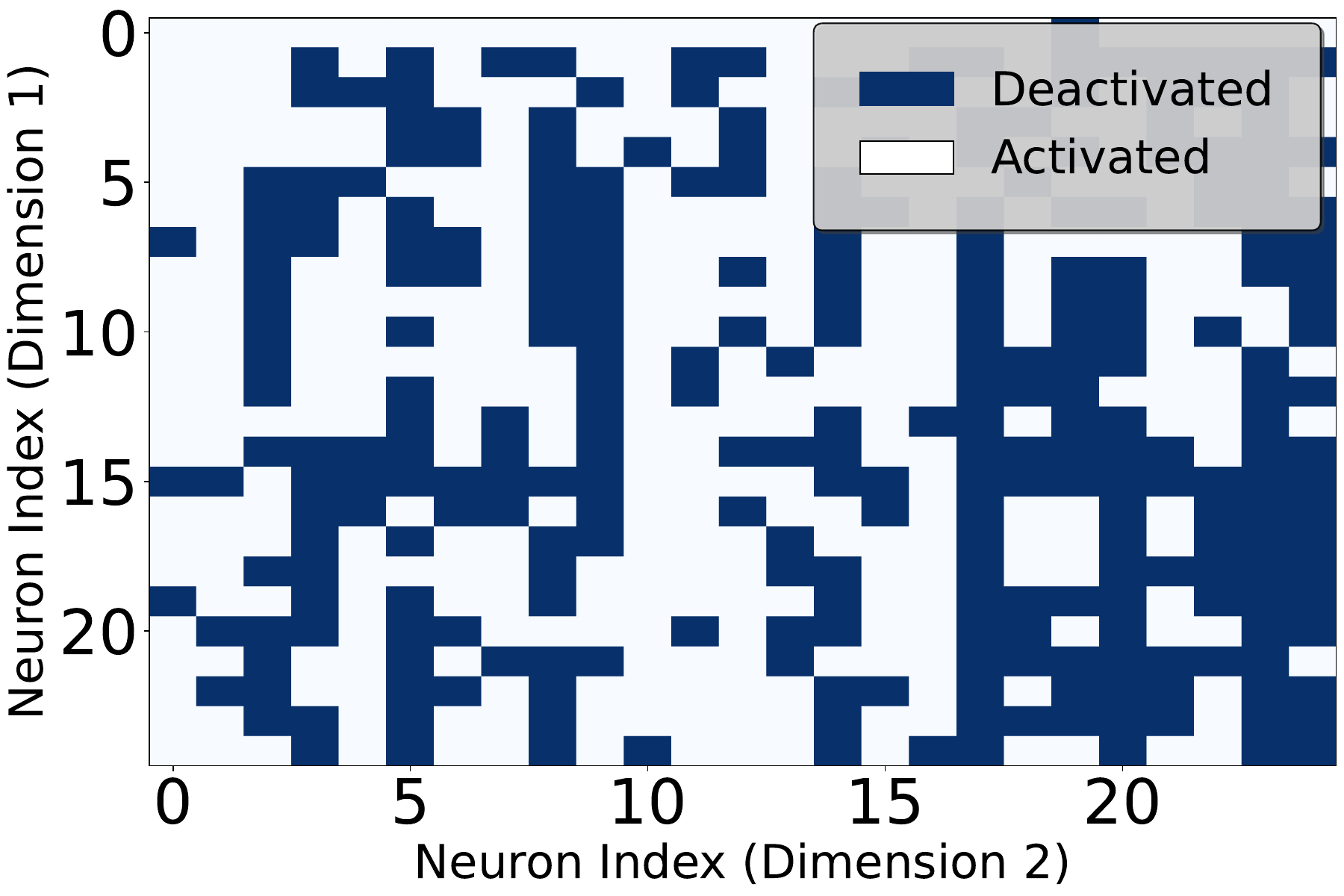}
         \caption{\centering Layer 10}
         \label{fig:Sample8_100_layer10}
     \end{subfigure}
     \hfill
     \begin{subfigure}[b]{0.24\textwidth}
         \centering
         \includegraphics[width=\textwidth]{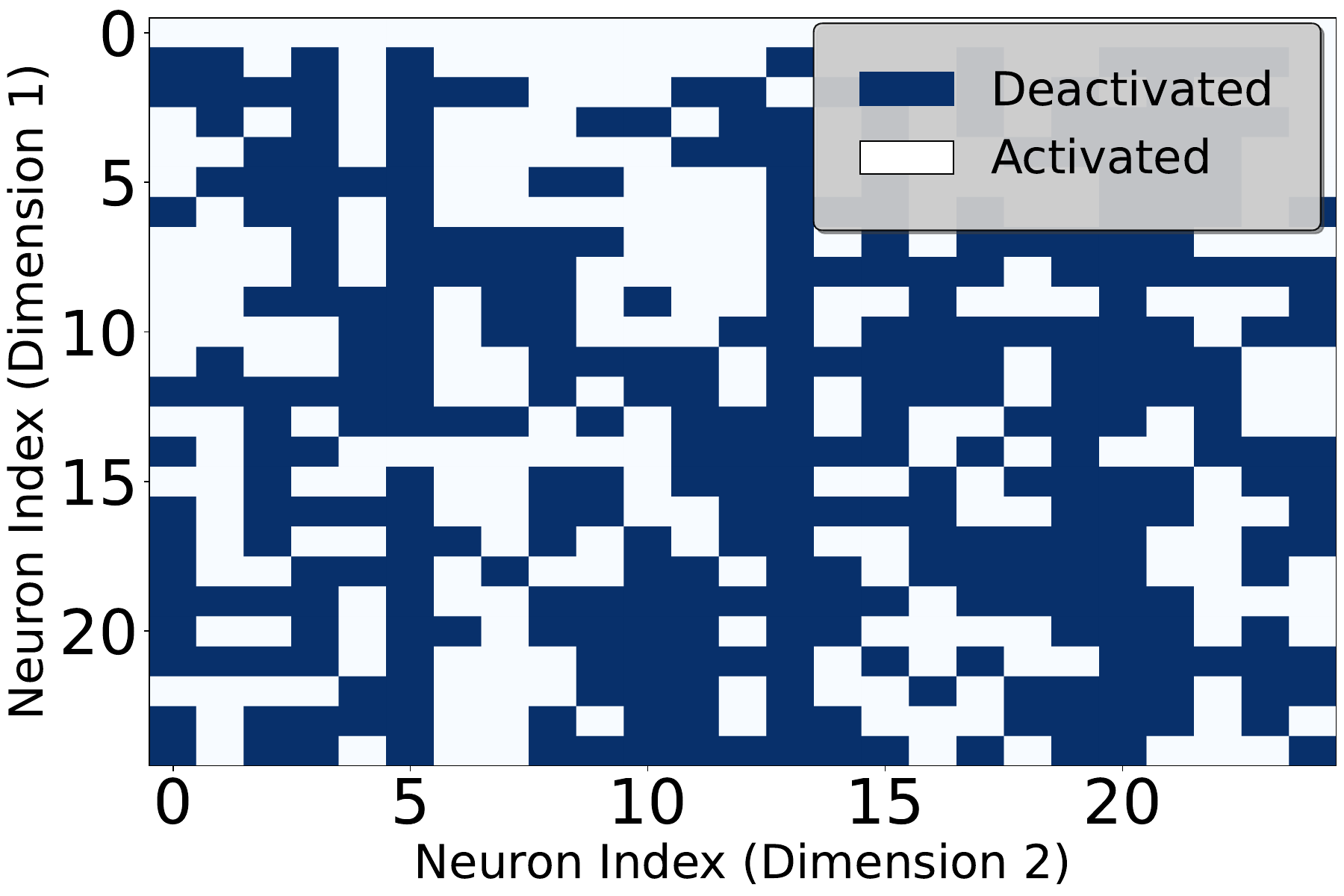}
         \caption{\centering Layer 20}
         \label{fig:Sample8_100_layer20}
     \end{subfigure}
     \hfill
     \begin{subfigure}[b]{0.24\textwidth}
         \centering
         \includegraphics[width=\textwidth]{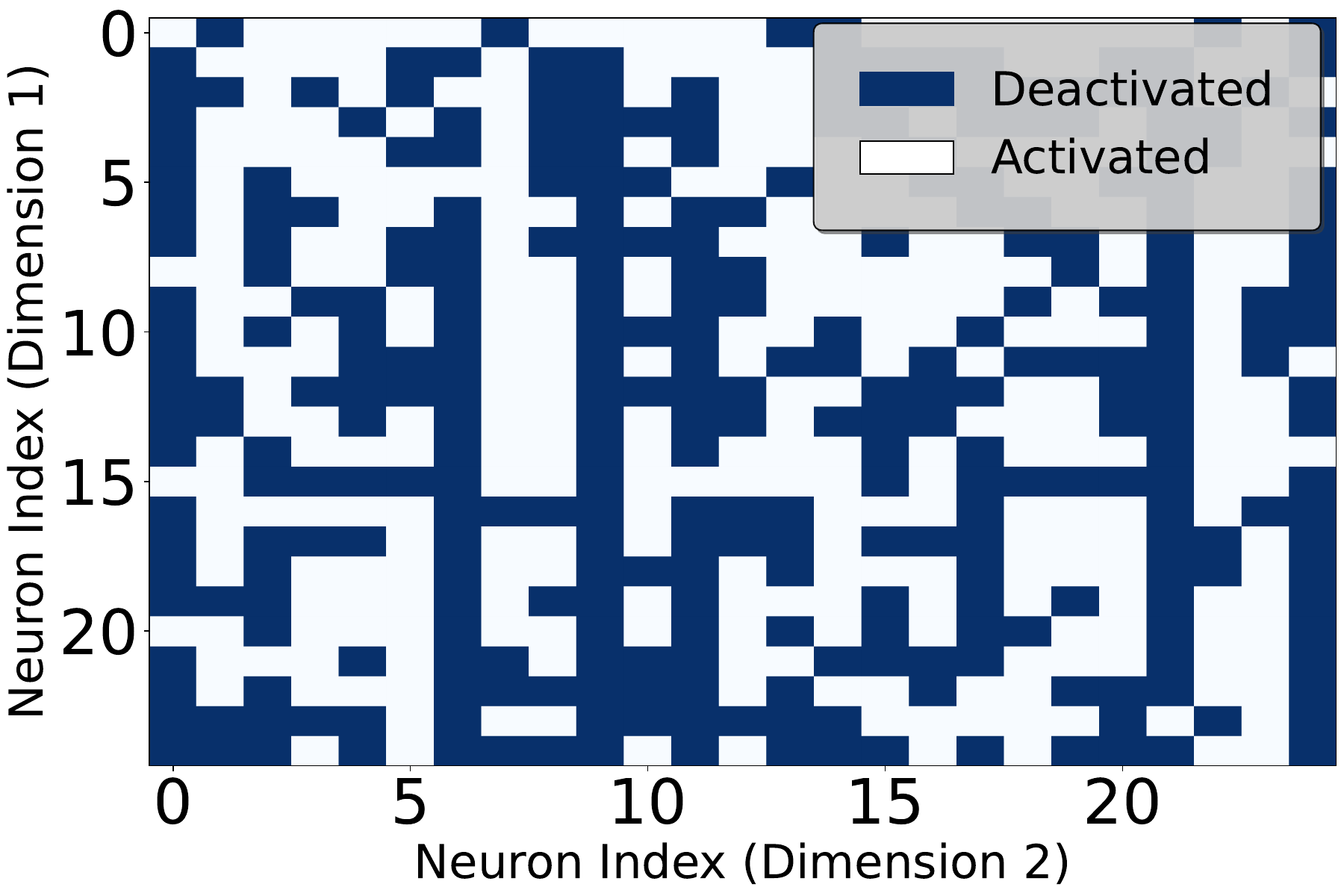}
         \caption{\centering Layer 32}
         \label{fig:Sample8_100_layer32}
     \end{subfigure}
        \vspace{-0.3em}
        \caption{Activation heatmap pattern of LLaMa-3-8b with default Sample 8 input}
        \label{fig:Sample_8_default}
        \vspace{-0.3em}
\end{figure*}

\begin{figure*}[ht]
     \centering
     \begin{subfigure}[b]{0.24\textwidth}
         \centering
         \includegraphics[width=\textwidth]{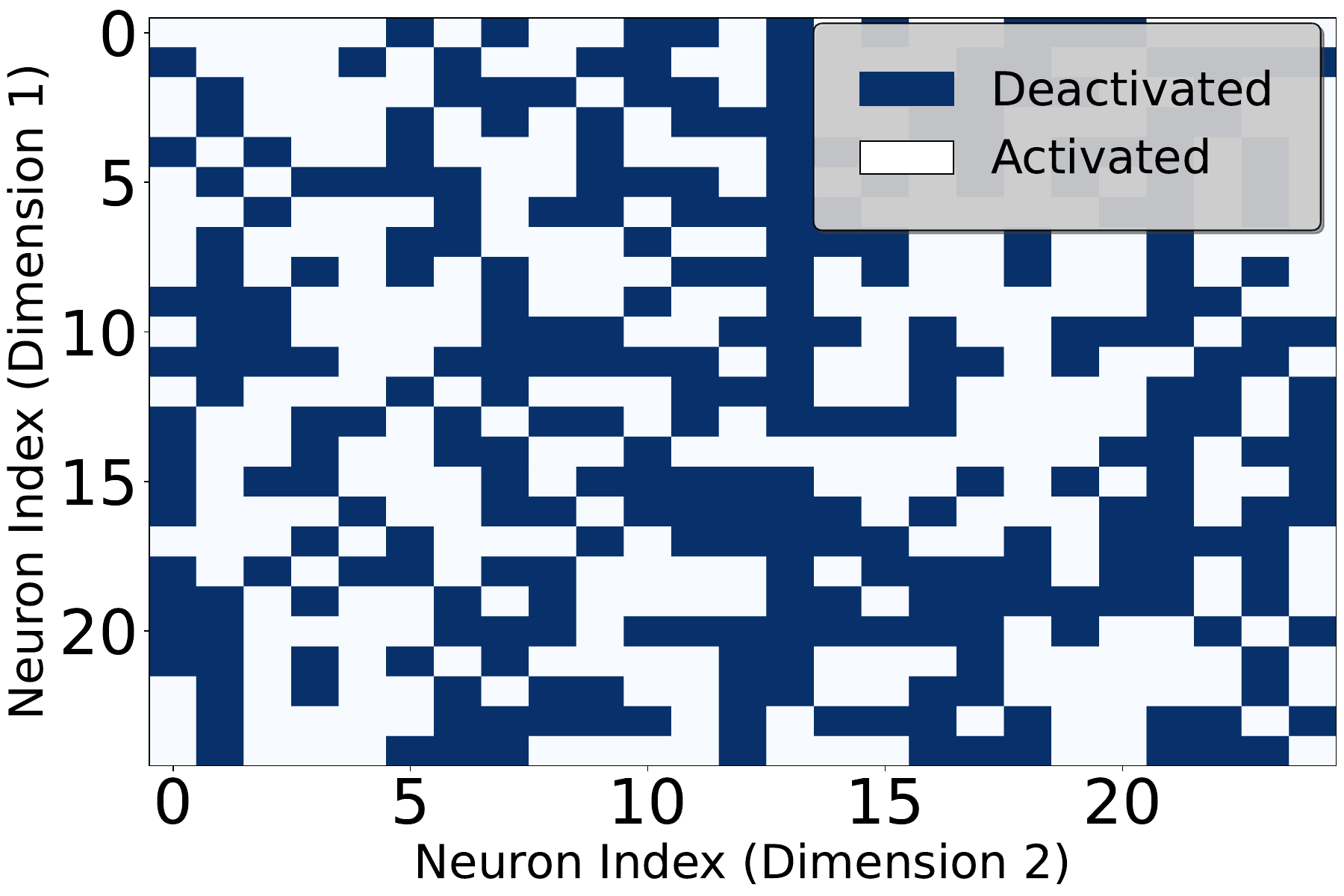}
         \caption{\centering Layer 1}
         \label{fig:Sample8_70_layer1}
     \end{subfigure}
     \hfill
     \begin{subfigure}[b]{0.24\textwidth}
         \centering
         \includegraphics[width=\textwidth]{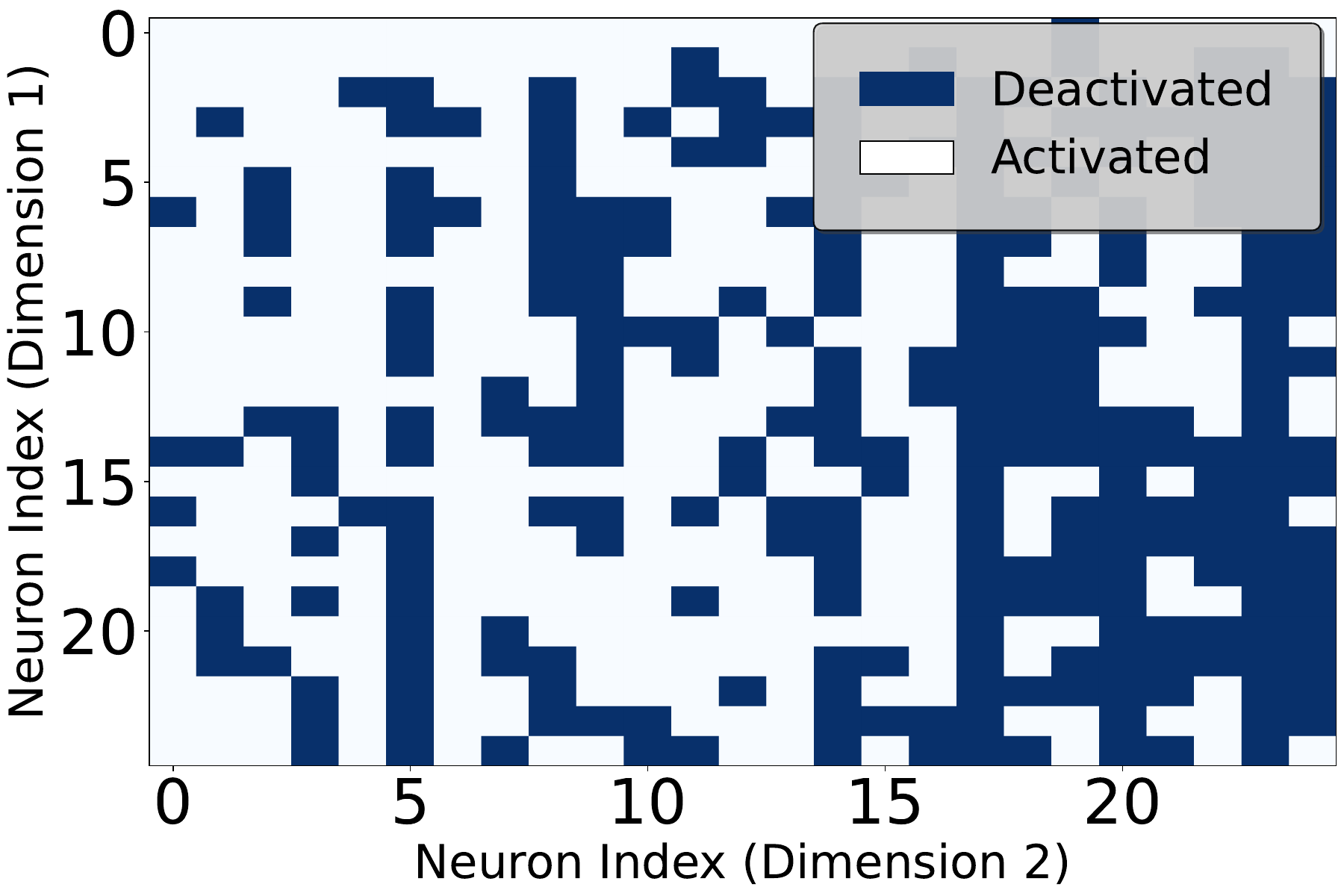}
         \caption{\centering Layer 10}
         \label{fig:Sample8_70_layer10}
     \end{subfigure}
     \hfill
     \begin{subfigure}[b]{0.24\textwidth}
         \centering
         \includegraphics[width=\textwidth]{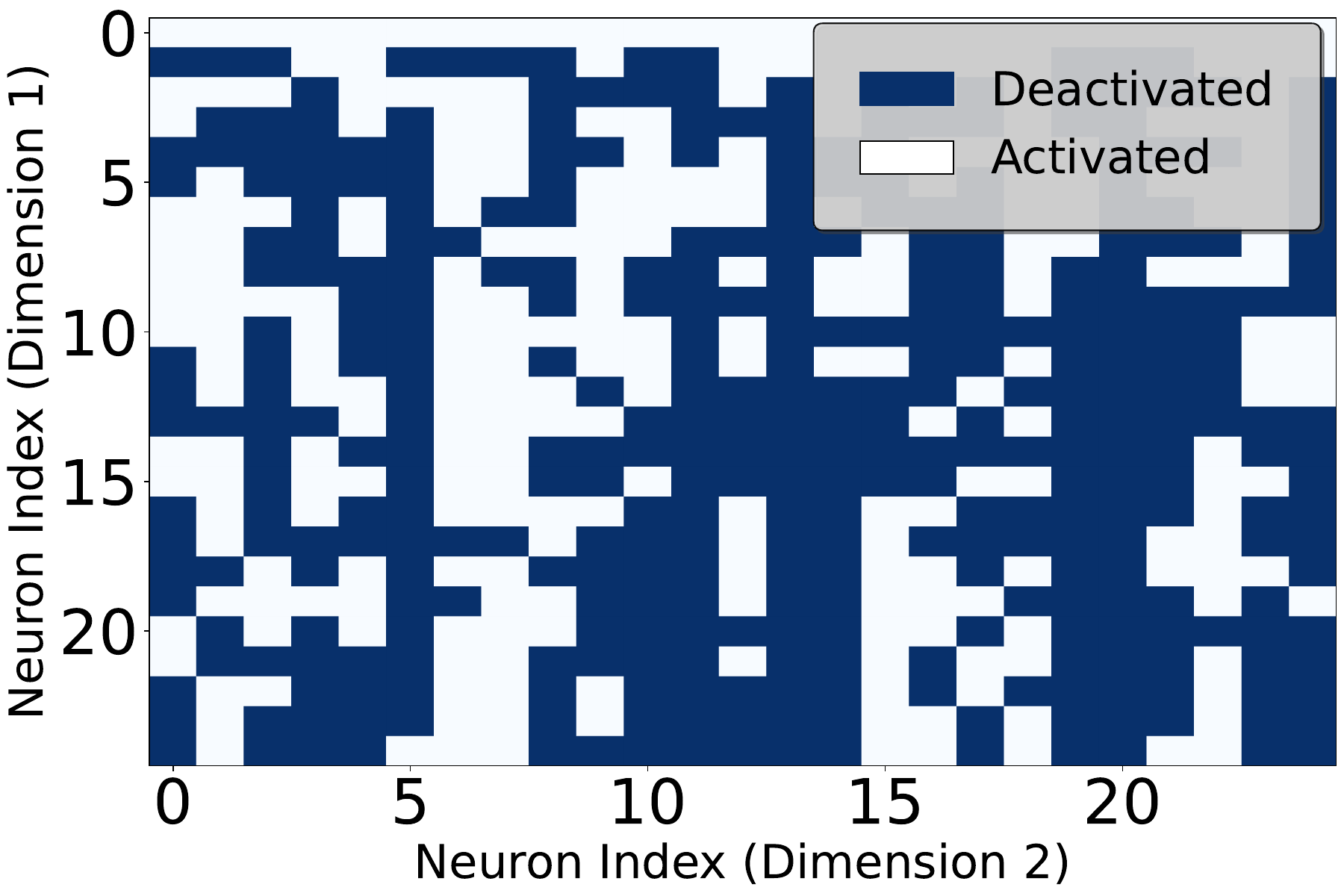}
         \caption{\centering Layer 20}
         \label{fig:Sample8_70_layer20}
     \end{subfigure}
     \hfill
     \begin{subfigure}[b]{0.24\textwidth}
         \centering
         \includegraphics[width=\textwidth]{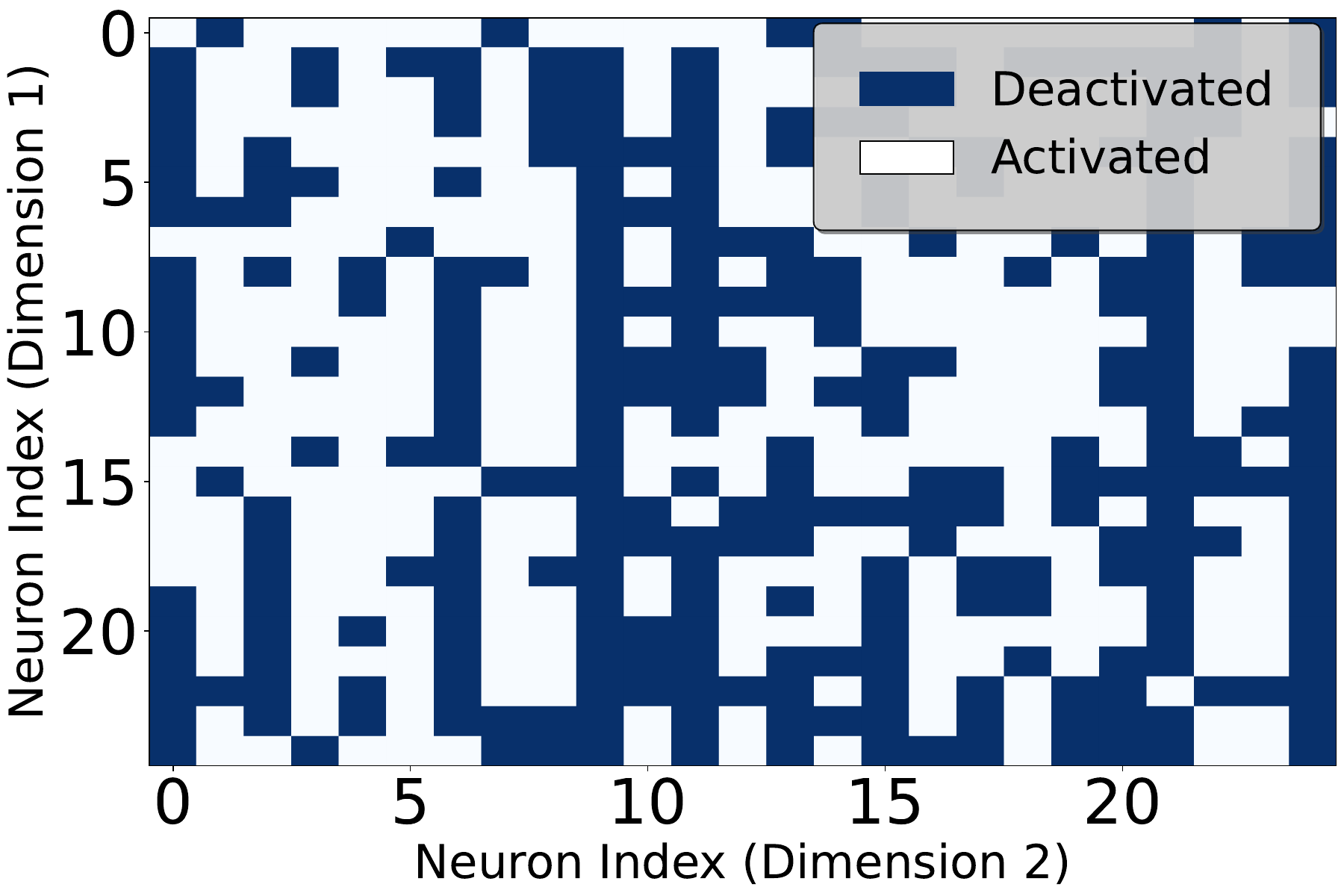}
         \caption{\centering Layer 32}
         \label{fig:Sample8_70_layer32}
     \end{subfigure}
        \vspace{-0.3em}
        \caption{\centering Activation heatmap pattern of LLaMa-3-8b with 70\% similarity Sample 8 input; Comparing with default Sample 8's patterns in Figure~\ref{fig:Sample_8_default}, matching rates are 57\%, 71.51\%, 71.51\%, and 64.06\% for Layer 1, 10, 20, and 32 , respectively.}
        \label{fig:Sample_8_70}
        \vspace{-0.3em}
\end{figure*}

\subsection{Potential Prediction Approach}

\noindent
LLM typically consists of multiple deep layers. E.g., \textit{LLaMA-3-8B} has 32 layers. During the LLM inference process, the earlier layer executes first, and its outputs are forwarded to the following layer for further processing. Therefore, we can use the earlier layer's activation patterns to predict future layers' activation patterns. E.g., we may use the pattern in \textit{Layer 1} to predict the patterns of \textit{Layer 16-32}. As shown in ShadowLLM's \cite{akhauri2024shadowllmpredictorbasedcontextualsparsity} approach, a lightweight predictor, with only one hidden layer, is capable of predicting activation patterns in future layers. This method introduces minimal computational overhead compared to the benefits it provides. Selectively prefetching the predicted weights significantly reduces memory and CPU usage during inference. The system has time to prefetch the predicted weights in the future layers and reduce the disk data loading time. From the main memory's perspective, the executing LLM is effectively compressed if we can achieve a reasonable success rate for pattern prediction.

To evaluate the predictability of the LLM activation patterns, we randomly select 12 samples as the input prompts, where each sample consists of 2048 tokens. User input prompts are always difficult to match 100\%, even when they want to ask the same question. Therefore, we also develop variants for these 12 samples with 70\% to 95\% similarity to tolerate the inconsistency of user inputs. The '70\% to 95\% similarity' indicates the '30\% to 5\%' modifications on the original sample. Considering many recent LLMs using the same SwiGLU activation function, for simplicity, we evaluate \textit{LLaMa-3-8b} as a representative in this section. 

\vspace{2pt}
\subsection{Pattern Similarity and Predictability Analysis}
\noindent
Table~\ref{tab:similarity} summarizes the activation pattern match rates of \textit{LLaMa-3-8B Layer 1} with different but similar input tokens. For each default input sample (total 12 samples), we developed 6 variants with 95\% to 70\% similarity. Even with 30\% input content change (70\% similarity), we still observe that the majority of evaluated samples (9/12) get 100\% match rates. For the worst similarity, we obtain 53\% activation match rate with 70\% input similarity. From Table~\ref{tab:similarity}, we conclude that the activation patterns are not as diverse as user inputs. Similar user inputs are highly possible to share the same activation pattern. This many-to-one mapping feature limits the total number of possible activation outputs, which is the essential prerequisite of effective pattern prediction. Also, Table~\ref{tab:similarity} illustrates the similarity results of \textit{LLaMa-3-8B Layer 1}. However, we should further confirm if the high matching rate of activation patterns would also be applied in the deeper LLM layers. If matching rates of all layers are high, then theoretically, we can design a predictor to prefetch associate activate neurons for all layers according to user inputs. 

Figure~\ref{fig:Sample_1_default} and Figure~\ref{fig:Sample_1_70} are the comparisons for \textit{Sample 1} where their input similarity is only 70\%. For easier reading, we just show the tiny piece of the activation heatmap ($25\times25$ neurons) in four layers. For \textit{Sample 1}, even using varying input prompts (70\% similarity), the matching rates of activation patterns of all layers are 100\%. Figure~\ref{fig:Sample_8_default} and Figure~\ref{fig:Sample_8_70} are the pattern comparisons for \textit{Sample 8} and its 70\% similarity variant. The matching percentages of \textit{Layer 1},  \textit{Layer 10}, \textit{Layer 20}, and \textit{Layer 32} are 57\%, 71.51\%, 71.51\%, and 64.06\%, respectively. An effective predictor with prefetching mechanisms may still help in this worst sample type because we can prefetch the majority of neurons in FFN components and, therefore, cut down data loading time from disk. 

\vspace{2pt}
\noindent \textbf{\underline{Insight 1}}: The results from Table~\ref{tab:similarity} indicate that LLM activation patterns are possible to share with many similar user inputs. This many-to-one mapping feature narrows the possible output categories and is an essential prerequisite for effective activation pattern prediction.

\vspace{2pt}
\noindent \textbf{\underline{Insight 2}}: Most evaluated input samples have 100\% same activation patterns for all layers with their 70\% similarity variants. The activation patterns of all LLM layers can be considered as an entire predicted output and selected based on \textit{Layer 1}'s activation pattern. In other words, the activation patterns and state-of-the-art LLMs are predictable and can effectively compress new large language models from the main memory's perspective.

\vspace{2pt}
\noindent \textbf{\underline{Insight 3}}:  Combining the tradeoff conclusion between sparsity and perplexity from Section~\ref{sec:ppl_sparsity}, the effective activation pattern prediction and prefetching can compress 50\% of LLMs from system resources' perspective. Therefore, it allows better LLM execution on resource-constraint edge devices.

\section{Related Works}
\label{sec:related}
\noindent\textit{\textbf{LLM Compression for Edge Systems.}} 
Despite the impressive growing capabilities of LLMs, the quick escalation in model size has led to an exponential increase in the memory and computational demands, presenting significant deployment challenges (Kaplan \textit{et al.}~\cite{Kaplan}; Liu \textit{et al.}~\cite{DejaVu}) in resource-constraint edge devices. Various compression techniques have been proposed to harvest the benefits of local LLM intelligence in edge devices, which includes  quantization~\cite{han2016deep, jacob2017quantization},  pruning ~\cite{han2016deep, molchanov2017pruning} and model distillation ~\cite{hinton2015distilling, tang2019distilling, gu2024minillm}. Additionally, efficient sampling methods~\cite{pmlr-v202-leviathan23a, Wang} have been proposed to expedite inference decoding. Our work investigates the potential of forced activation sparsity, which is orthogonal to the above-mentioned approaches and may combine with them to enhance the LLM compression further.

\vspace{3pt}
\noindent\textit{\textbf{Activation Sparsity Utilization.}} 
Activation sparsity occurs when a portion of activation neuron outputs are zero. The computation of these inactive neurons can be ignored, which indicates that they are unnecessary to be fetched into the main memory and execute the computation. Li \textit{et al.}~\cite{li2023lazy} and Liu \textit{et al.}~\cite{DejaVu} have recognized the intrinsic activation sparsity found in several LLMs and have utilized this characteristic to speed up inference. These techniques can be combined to enhance performance effectively. Despite these insights, no prior work has provided concrete model results after employing activation sparsity. However, our study aims to bridge this gap by not only implementing activation sparsity but also presenting detailed performance metrics, thus validating the practical benefits of this approach.

\vspace{3pt}
\noindent
\textit{\textbf{ReLUfication.}} 
The activation sparsity benefits from previous research are limited to ReLU-based Transformer architectures, including LLMs such as T5 (Raffel \textit{et al.}~\cite{raffel2023exploring}) and OPT (Zhang \textit{et al.}~\cite{zhang2022opt}), as well as in some vision models like ViT (Dosovitskiy \textit{et al.}~\cite{dosovitskiy2021image}). However, recent LLMs such as Falcon (Almazrouei \textit{et al.}~\cite{almazrouei2023falcon}) and LLaMa (Touvron \textit{et al.}~\cite{touvron2023llama}) predominantly use non-ReLU activation functions like GELU (Hendrycks and Gimpel \textit{et al.}~\cite{hendrycks2023gaussian}) and Swish (Elfwing \textit{et al.}~\cite{elfwing2017sigmoidweighted}), which do not directly provide activation sparsity. To utilize the benefits of activation sparsity without developing a ReLU-activated LLM from scratch, many researchers have adopted a technique known as ReLUfication. This method introduces sparse ReLU-based activations into non-ReLU LLMs. For instance, Zhang \textit{et al.}~\cite{zhang-etal-2022-moefication} successfully converted a GELU-activated BERT (Devlin \textit{et al.}~\cite{devlin2019bert}) into a ReLU-activated version by directly swapping the activation function. Similarly, Zhang \textit{et al.}~\cite{zhang2024relu2} applied this approach to Falcon and LLaMa, which originally used GELU and Swish, respectively. However, simply replacing the activation function often fails to achieve satisfactory sparsity, primarily due to the unaddressed inherent limitations of the original non-ReLU activation distributions. Instead of utilizing ReLUfication, we propose to enforce sparsity without altering the activation function. This method retains the original activation function's properties while leveraging sparsity benefits. Our approach directly omits the lower magnitude values to induce sparsity, offering a novel and effective method for LLM compression.

\section{Conclusion}
\label{sec:concl}
\noindent
Deploying LLM on edge devices offers numerous compelling benefits. Recently, major tech companies have started to invest in and seek edge LLM solutions. However, the insufficient computing and memory resources on edge devices restricted effective LLM executions. Most regular edge devices are still unable to meet the requirements of LLM execution, even when combined with existing compression techniques. Therefore, we explore new schemes that could seamlessly combine with regular LLM compression approaches. The new activation functions of state-of-the-art LLMs eliminate the feasibility of Apple's compression approach, which performs well for ReLU-based LLMs. In this work, we search for new compression opportunities from activation sparsity, which can benefit the majority of general LLMs and not be bound by specific activation functions. Our systematic explorations indicate that we can safely secure 50\% extra sparsity in FFN layers with a negligible accuracy loss for the state-of-the-art LLMs. This finding allows us to compress the LLMs from memory's perspective via prediction and prefetching. Moreover, to verify the predictability of LLM activation patterns, we also analyze matching rates of LLM activation patterns with multiple user input variants. We finally conclude that the LLM activation patterns are highly predictable and have a huge potential to get extra sparsity. The paper provides a guideline for design pattern predictors with prefetching capabilities, leading to less LLM execution latency, lower power cost, and improved user experience. Moreover, besides general LLMs, our new compression approach can be simply extended to other large-scale transformer-based AI models.  

\section*{Acknowledgement} 
We are grateful to the anonymous reviewers for their comments and suggestions on this paper. This work was supported in part by U.S. National Science Foundation (NSF) grants CPS-2103459, SHF-2210744.


\bibliographystyle{IEEEtran}
\bibliography{IEEEexample.bib}

\begin{thebibliography}{10}
\providecommand{\url}[1]{#1}
\csname url@samestyle\endcsname
\providecommand{\newblock}{\relax}
\providecommand{\bibinfo}[2]{#2}
\providecommand{\BIBentrySTDinterwordspacing}{\spaceskip=0pt\relax}
\providecommand{\BIBentryALTinterwordstretchfactor}{4}
\providecommand{\BIBentryALTinterwordspacing}{\spaceskip=\fontdimen2\font plus
\BIBentryALTinterwordstretchfactor\fontdimen3\font minus \fontdimen4\font\relax}
\providecommand{\BIBforeignlanguage}[2]{{%
\expandafter\ifx\csname l@#1\endcsname\relax
\typeout{** WARNING: IEEEtran.bst: No hyphenation pattern has been}%
\typeout{** loaded for the language `#1'. Using the pattern for}%
\typeout{** the default language instead.}%
\else
\language=\csname l@#1\endcsname
\fi
#2}}
\providecommand{\BIBdecl}{\relax}
\BIBdecl

\bibitem{lubbad2023gpt}
M.~Lubbad, ``Gpt-4 parameters: Unlimited guide nlp’s game-changer,'' 2023.

\bibitem{ndhar}
N.~Dhar, B.~Deng, D.~Lo, X.~Wu, L.~Zhao, and K.~Suo, ``An empirical analysis and resource footprint study of deploying large language models on edge devices,'' 2024.

\bibitem{ullah2024role}
A.~Ullah, G.~Qi, S.~Hussain, I.~Ullah, and Z.~Ali, ``The role of llms in sustainable smart cities: Applications, challenges, and future directions,'' 2024.

\bibitem{yin2024llm}
W.~Yin, M.~Xu, Y.~Li, and X.~Liu, ``Llm as a system service on mobile devices,'' 2024.

\bibitem{pruning}
J.~Yu, A.~Lukefahr, and Palframan..., ``Scalpel: Customizing dnn pruning to the underlying hardware parallelism,'' 2017.

\bibitem{Yang_2019_CVPR}
J.~Yang, X.~Shen, J.~Xing, X.~Tian, H.~Li, B.~Deng, J.~Huang, and X.-s. Hua, ``Quantization networks,'' 2019.

\bibitem{gou2021knowledge}
J.~Gou, B.~Yu, S.~J. Maybank, and D.~Tao, ``Knowledge distillation: A survey,'' \emph{International Journal of Computer Vision}, 2021.

\bibitem{kepner2020graphchallenge}
J.~Kepner, S.~Alford, V.~Gadepally, M.~Jones, L.~Milechin, A.~Reuther, R.~Robinett, and S.~Samsi, ``Graphchallenge. org sparse deep neural network performance,'' \emph{arXiv preprint arXiv:2004.01181}, 2020.

\bibitem{DejaVu}
Z.~Liu, J.~Wang, T.~Dao, T.~Zhou, B.~Yuan, Z.~Song, A.~Shrivastava, C.~Zhang, Y.~Tian, C.~R\'{e}, and B.~Chen, ``Deja vu: contextual sparsity for efficient llms at inference time.''\hskip 1em plus 0.5em minus 0.4em\relax JMLR.org, 2023.

\bibitem{alizadeh2024llm}
K.~Alizadeh, I.~Mirzadeh, D.~Belenko, K.~Khatamifard, M.~Cho, C.~C.~D. Mundo, M.~Rastegari, and M.~Farajtabar, ``Llm in a flash: Efficient large language model inference with limited memory,'' 2024.

\bibitem{liu2022fl}
J.~Liu, Y.~Song, K.~Xue, H.~Sun, C.~Wang, L.~Chen, H.~Jiang, J.~Liang, and T.~Ruan, ``Fl-tuning: Layer tuning for feed-forward network in transformer,'' \emph{arXiv preprint arXiv:2206.15312}, 2022.

\bibitem{wikitext}
S.~Merity, C.~Xiong, J.~Bradbury, and R.~Socher, ``Pointer sentinel mixture models,'' 2016.

\bibitem{pires2023wide}
T.~P. Pires, A.~V. Lopes, Y.~Assogba, and H.~Setiawan, ``One wide feedforward is all you need,'' 2023.

\bibitem{akhauri2024shadowllmpredictorbasedcontextualsparsity}
\BIBentryALTinterwordspacing
Y.~Akhauri, A.~F. AbouElhamayed, J.~Dotzel, Z.~Zhang, A.~M. Rush, S.~Huda, and M.~S. Abdelfattah, ``Shadowllm: Predictor-based contextual sparsity for large language models,'' 2024. [Online]. Available: \url{https://arxiv.org/abs/2406.16635}
\BIBentrySTDinterwordspacing

\bibitem{Kaplan}
J.~Kaplan, S.~McCandlish, T.~Henighan, T.~B. Brown, B.~Chess, R.~Child, S.~Gray, A.~Radford, J.~Wu, and D.~Amodei, ``Scaling laws for neural language models,'' 2020.

\bibitem{han2016deep}
S.~Han, H.~Mao, and W.~J. Dally, ``Deep compression: Compressing deep neural networks with pruning, trained quantization and huffman coding,'' 2016.

\bibitem{jacob2017quantization}
B.~Jacob, S.~Kligys, B.~Chen, M.~Zhu, M.~Tang, A.~Howard, H.~Adam, and D.~Kalenichenko, ``Quantization and training of neural networks for efficient integer-arithmetic-only inference,'' 2017.

\bibitem{molchanov2017pruning}
P.~Molchanov, S.~Tyree, T.~Karras, T.~Aila, and J.~Kautz, ``Pruning convolutional neural networks for resource efficient inference,'' 2017.

\bibitem{hinton2015distilling}
G.~Hinton, O.~Vinyals, and J.~Dean, ``Distilling the knowledge in a neural network,'' 2015.

\bibitem{tang2019distilling}
R.~Tang, Y.~Lu, L.~Liu, L.~Mou, O.~Vechtomova, and J.~Lin, ``Distilling task-specific knowledge from bert into simple neural networks,'' 2019.

\bibitem{gu2024minillm}
Y.~Gu, L.~Dong, F.~Wei, and M.~Huang, ``Minillm: Knowledge distillation of large language models,'' 2024.

\bibitem{pmlr-v202-leviathan23a}
Y.~Leviathan, M.~Kalman, and Y.~Matias, ``Fast inference from transformers via speculative decoding.''\hskip 1em plus 0.5em minus 0.4em\relax PMLR, 2023.

\bibitem{Wang}
Y.~Wang, K.~Chen, H.~Tan, and K.~Guo, ``Tabi: An efficient multi-level inference system for large language models.''\hskip 1em plus 0.5em minus 0.4em\relax Association for Computing Machinery, 2023.

\bibitem{li2023lazy}
Z.~Li, C.~You, S.~Bhojanapalli, D.~Li, A.~S. Rawat, S.~J. Reddi, K.~Ye, F.~Chern, F.~Yu, R.~Guo, and S.~Kumar, ``The lazy neuron phenomenon: On emergence of activation sparsity in transformers,'' 2023.

\bibitem{raffel2023exploring}
C.~Raffel, N.~Shazeer, A.~Roberts, K.~Lee, S.~Narang, M.~Matena, Y.~Zhou, W.~Li, and P.~J. Liu, ``Exploring the limits of transfer learning with a unified text-to-text transformer,'' 2023.

\bibitem{zhang2022opt}
S.~Zhang, S.~Roller, N.~Goyal, M.~Artetxe, M.~Chen, S.~Chen, C.~Dewan, M.~Diab, X.~Li, X.~V. Lin, T.~Mihaylov, M.~Ott, S.~Shleifer, K.~Shuster, D.~Simig, P.~S. Koura, A.~Sridhar, T.~Wang, and L.~Zettlemoyer, ``Opt: Open pre-trained transformer language models,'' 2022.

\bibitem{dosovitskiy2021image}
A.~Dosovitskiy, L.~Beyer, A.~Kolesnikov, D.~Weissenborn, X.~Zhai, T.~Unterthiner, M.~Dehghani, M.~Minderer, G.~Heigold, S.~Gelly, J.~Uszkoreit, and N.~Houlsby, ``An image is worth 16x16 words: Transformers for image recognition at scale,'' 2021.

\bibitem{almazrouei2023falcon}
E.~Almazrouei, H.~Alobeidli, A.~Alshamsi, A.~Cappelli, R.~Cojocaru, M.~Debbah, Étienne Goffinet, D.~Hesslow, J.~Launay, Q.~Malartic, D.~Mazzotta, B.~Noune, B.~Pannier, and G.~Penedo, ``The falcon series of open language models,'' 2023.

\bibitem{touvron2023llama}
H.~Touvron, L.~Martin, K.~Stone, and e.~a. Peter~Albert, ``Llama 2: Open foundation and fine-tuned chat models,'' 2023.

\bibitem{hendrycks2023gaussian}
D.~Hendrycks and K.~Gimpel, ``Gaussian error linear units (gelus),'' 2023.

\bibitem{elfwing2017sigmoidweighted}
S.~Elfwing, E.~Uchibe, and K.~Doya, ``Sigmoid-weighted linear units for neural network function approximation in reinforcement learning,'' 2017.

\bibitem{zhang-etal-2022-moefication}
Z.~Zhang, Y.~Lin, Z.~Liu, P.~Li, M.~Sun, and J.~Zhou, ``{M}o{E}fication: Transformer feed-forward layers are mixtures of experts.''\hskip 1em plus 0.5em minus 0.4em\relax Association for Computational Linguistics, 2022.

\bibitem{devlin2019bert}
J.~Devlin, M.-W. Chang, K.~Lee, and K.~Toutanova, ``Bert: Pre-training of deep bidirectional transformers for language understanding,'' 2019.

\bibitem{zhang2024relu2}
Z.~Zhang, Y.~Song, G.~Yu, X.~Han, Y.~Lin, C.~Xiao, C.~Song, Z.~Liu, Z.~Mi, and M.~Sun, ``Relu$^2$ wins: Discovering efficient activation functions for sparse llms,'' \emph{arXiv preprint arXiv:2402.03804}, 2024.

\end{thebibliography}

\end{document}